\documentclass[acmtocl,acmnow]{acmtrans2m}

\usepackage{epsfig}

\newtheorem{theorem}{Theorem}[section]

\newtheorem{corollary}[theorem]{Corollary}
\newtheorem{proposition}[theorem]{Proposition}
\newtheorem{lemma}[theorem]{Lemma}
\newdef{definition}[theorem]{Definition}
\newdef{remark}[theorem]{Remark}
\newtheorem{observation}{Observation}[section]
\newtheorem{example}{Example}
\newtheorem{algorithm}{Algorithm}[section]

\long\def\COMMENT#1\ENDCOMMENT{\message{(Commented text...)}\par}

\def\st{\smallskip \noindent}

\def\ni{\noindent}
\def\and{ \ \wedge}

\def\beq{\begin{equation}}
\def\eeq#1{\label{#1}\end{equation}}

\def\calS0{{\cal S}_0}

\def\qed{{\hfill$\Box$}}

\title{Regression with respect to sensing actions and partial
states}

\author{LE-CHI TUAN, CHITTA BARAL\\
Arizona State University \\ and\\
TRAN CAO SON \\
New Mexico State University}

\begin{abstract}
In this paper, we present a \emph{state-based regression function}
for planning domains where an agent does not have complete
information and may have sensing actions. We consider  binary
domains and employ the 0-approximation [Son \& Baral 2001] to
define the regression function. In binary domains, the use of
0-approximation means using 3-valued states. Although planning
using this approach is incomplete with respect to the full
semantics, we adopt it to have a lower complexity. We prove the
soundness and completeness of our regression formulation with
respect to the definition of progression. More specifically, we
show that (i) a plan obtained through regression for a planning
problem is indeed a progression solution of that planning problem,
and that (ii) for each plan found through progression, using
regression one obtains that plan or an equivalent one. We then
develop a conditional planner that utilizes our regression
function. We prove the soundness and completeness of our planning
algorithm and present experimental results with respect to several
well known planning problems in the literature.
\end{abstract}





\begin{document}

\begin{bottomstuff}
Author's address: Le-chi Tuan, Chitta Baral, Computer Science and
Engineering, Arizona State University, Tempe, AZ 85287, USA. {\it
\{lctuan,baral\}@asu.edu} \newline Tran Cao Son, Computer Science
Department, New Mexico State University, Las Cruces, NM 88003,
USA. {\it tson@cs.nmsu.edu}\newline This is a revised and extended
version of a paper accepted to The Nineteenth National Conference
on Artificial Intelligence (AAAI'04), San Jose, USA.
\end{bottomstuff}

\maketitle

\section{Introduction and Motivation}

\subsection{Introduction and Motivation}

An important aspect in reasoning about actions and in
characterizing the semantics of action description languages is to
define a transition function encoding the transition between
states due to actions. This transition function is often viewed as
a {\em progression} function in that it denotes the progression of
the world by the execution of actions. The `opposite' or `inverse'
of progression is referred to as {\em regression}.

\st Even for the simple case where we have only non-sensing
actions and the progression transition function is deterministic,
there are various formulations of regression. For example,
consider the following. Let $\Phi$ be the progression transition
function from actions and states to states. I.e., intuitively,
$\Phi(a,s) = s'$ means that if the action $a$ is executed in state
$s$ then the resulting state will be $s'$. One way to define a
regression function $\Psi_1$ is to define it with respect to
states. In that case $s \in \Psi_1(a,s')$ will mean that the state
$s'$ is reached if $a$ is executed in $s$. Another way regression
is defined is with respect to formulas. In that case $\Psi_2(a,f)
= g$, where $f$ and $g$ are formulas, means that if $a$ is
executed in a state satisfying $g$ then a state satisfying $f$
will be reached.

\st For planning using heuristic search, often a different
formulation of regression is given. Since most planning research
is about goals that are conjunction of literals, regression is
defined with respect to a set of literals and an action. In that
case the conjunction of literals (often specifying the goal)
denotes a set of states, one of which needs to be reached. This
regression is slightly different from $\Psi_2$ as the intention is
to regress to another set of literals (not an arbitrary formula),
denoting a sub-goal.

\st With respect to the planning language STRIPS, where each
action $a$ has an add list $Add(a)$, a delete list $Del(a)$, and a
precondition list $Prec(a)$, the progression function is defined
as $Progress(s,a) = s + Add(a) - Del(a)$; and the regression
function is defined as $Regress(conj,a) = conj + Prec(a) -
Add(a)$, where $conj$ is a set of atoms. The relation between
these two, formally proven in \cite{ped86}, shows the correctness
of regression based planners; which in recent years through use of
heuristics (e.g. \cite{BG01,nguyenetal02}) have done exceedingly
well on planning competitions.

\st {\em In this paper we are concerned with domains where the
agent does not have complete information about the world, and may
have sensing actions, which when executed do not change the world,
but rather give certain information about the world to the agent.}
As a result, plans may now no longer be simply a sequence of
(non-sensing) actions but may include sensing actions and
conditionals. Various formalisms have been developed for such
cases (e.g. \cite{L98,SB01}) and progression functions have been
defined. Also, the complexity of planning in such cases has been
analyzed in \cite{BKT99}. One approach to planning in the presence
of incomplete information is conformant planning where no sensing
action is used, and a plan is a sequence of actions leading to the
goal from every possible initial situation. However, this approach
proves inadequate for many planning problems \cite{SB01}, i.e.,
there are situations where sensing actions are necessary. In that
case, one approach is to use belief states or Kripke models
instead of states. It is shown that the total number of belief
states is double exponential while the total number of 3-valued
states is exponential in the number of fluents \cite{BKT99}. Here,
we pursue a provably less complex formulation with sensing actions
and use 3-valued states. In this approach, we will miss certain
plans, but that is the price we are willing to pay for reduced
complexity. This is consistent with and similar to the
considerations behind conformant planning. With that tradeoff in
mind, {\em in this paper we consider the 0-approximation semantics
defined in \cite{SB01} and define regression with respect to that
semantics.} We then formally relate our definition of regression
with the earlier definition of progression in \cite{SB01} and show
that planning using our regression function will not only give us
correct plans but also will not miss plans. We then use our
regression function in planning with sensing actions and show
that, even without using any heuristics, our planner produces very
good results.  To simplify our formulation, we only consider
STRIPS like actions where no
conditional effects are allowed. 


\st In summary the main contributions of our paper are:

\begin{itemize} 

\item [$\bullet$] A state-based regression function corresponding
to the 0-approximation semantics in \cite{SB01};

\item [$\bullet$] A formal result showing the soundness of our
regression function with respect to the progression transition
function in \cite{SB01};

\item [$\bullet$] A formal result showing the completeness of our
regression function with respect to the progression transition
function in \cite{SB01};


\item [$\bullet$] An algorithm that uses these regression
functions to construct conditional plans with sensing actions;

\item [$\bullet$] Implementation of this algorithm; and

\item [$\bullet$] Illustration of the performance of this
algorithm with respect to several examples in the literature.


\end{itemize}

\subsection{Related Work}

Our work in this paper is related to different approaches to
regression and planning in the presence of sensing actions and
incomplete information. It differs from earlier notion of
regression such as \cite{Rei98,SB01} in that our definition is
with respect to states while the earlier definitions are with
respect to formulas.

\st In the planning literature there has been a lot of work
\cite{PS92,CRT98,Eetal92,L98,STB03,SGP98,BG00,PC96,R00,R02,eit00}
in developing planners that generate conditional plans in presence
of incomplete information, some of which use sensing actions and
the others do not. Unlike the conditional planners
\cite{PS92,CRT98}, our planner can deal with sensing actions
similar to the planners in \cite{Eetal92,L98,STB03,SGP98}.
However, it does not deal with nondeterministic and probabilistic
actions such as the planners in \cite{BG00,PC96,R00,R02}. It is
also not a conformant planner as in \cite{CRT98,eit00}. For these
reasons, we currently compare our planner with those of
\cite{STB03,SGP98}.

\section{Background: 0-Approximation Semantics For A STRIPS-like Language}

\subsection{Action and Plan Representation} \label{pre}

We employ a STRIPS-like action representation~\cite{FN71} and
represent a planning problem by a tuple $P= \langle A,O,I,G
\rangle$ where $A$ is a finite set of fluents, $O$ is a finite set
of actions, and $I$ and $G$ are sets of fluent literals
\footnote{A fluent literal is either a positive fluent $f \in A$
or its negation (negative fluent) $\neg f$.} made up of fluents in
$A$. Intuitively, $I$ encodes what is known about the initial
state and $G$ encodes what is desired of a goal state.

\st An action $a \in O $ is either a {\em non-sensing action} or a
{\em sensing action} and is specified as follows:
\begin{itemize} 
    \item [$\bullet$]
    A non-sensing action $a$ is specified by an expression of the form

    \hspace{.2in} action $a$ \hspace{.2in} \parbox[t]{4.in}{:Pre $Pre_a$\\ :Add $Add_a$\\ :Del $Del_a$}

    where $Pre_a$ is a set of fluent literals representing the
    precondition for $a$'s execution, $Add_a$  and
    $Del_a$ are two disjoint sets of positive fluents representing
    the positive and negative effects of $a$, respectively; and

    \item [$\bullet$] A sensing action $a$ is specified by an expression of the form

    \hspace{.2in} action $a$ \hspace{.2in} \parbox[t]{4.in}{:Pre $Pre_a$\\ :Sense $Sens_a$}

    where $Pre_a$ is a set of fluent literals and $Sens_a$ is a set of
    positive fluents that do not appear in $Pre_a$.
\end{itemize}

\st To illustrate the action representation and our search
algorithm, we will use a small example, a version of the ``Getting
to Evanston'' from~\cite{SGP98}. Figure (\ref{example-fig01}) shows the actions of
this domain.

\begin{figure}[h]
\centerline{
\scriptsize
\begin{tabular}{|l|l|l|l|} \hline \hline
\multicolumn{1}{|c|} {{\bf Non-sensing action: Name}} &
\multicolumn{1}{|c|} {{\bf :Pre}} & \multicolumn{1}{|c|} {{\bf
:Add}} & \multicolumn{1}{|c|} {{\bf :Del}} \\ \hline \hline
goto-western-at-belmont & \{at-start\} & \{on-western, on-belmont\} & \{at-start\}\\
take-belmont & \{on-belmont, traffic-bad\} & \{on-ashland\} & \{on-western\}\\
take-ashland & \{on-ashland\} & \{at-evanston\} & \\
take-western & \{$\neg$traffic-bad, on-western\} & \{at-evanston\} & \\
\hline \hline \multicolumn{1}{|c|} {{\bf Sensing action: Name}} &
\multicolumn{1}{|c|} {{\bf :Pre}} & \multicolumn{2}{|c|} {{\bf
:Sense}} \\ \hline \hline
\multicolumn{1}{|c|} {check-traffic} & \multicolumn{1}{|c|} {$\emptyset$} & \multicolumn{2}{|c|} {\{traffic-bad\}} \\
\multicolumn{1}{|c|} {check-on-western} & \multicolumn{1}{|c|}
{$\emptyset$} & \multicolumn{2}{|c|} {\{on-belmont\}} \\ \hline
\end{tabular}
}
\normalsize
\caption{Actions of the ``Getting to Evanston'' domain.}
    \label{example-fig01}
\end{figure}

\st The notion of a plan in the presence of incomplete information
and sensing actions has been extensively discussed in the
literature~\cite{SL03,SB01}. In this paper, we consider {\em
conditional plans} that are formally defined as follows.
%
\begin {definition} [Conditional Plan] {\ } \label{defC}
\begin{itemize}
    \item [$\bullet$] An empty sequence of actions, denoted by $[\ ]$, is a conditional plan.
    \item [$\bullet$] If $a$ is a non-sensing action, then $a$ is a conditional plan.
    \item [$\bullet$] If $a$ is a sensing action,
    $\varphi_1,\ldots,\varphi_n$ are mutually exclusive
    conjunctions of fluent literals 
    , and $c_1, \ldots, c_n$ are conditional plans, then

        \centerline{$a; case(\varphi_1 \rightarrow c_1, \ldots, \varphi_n \rightarrow c_n)$}

        is a conditional plan \footnote{We often refer to this type of conditional plan as \emph{case plan
        }.}.
    \item [$\bullet$] if $c_1, c_2$ are conditional plans, then $c_1;c_2$ is a
        conditional plan.
    \item [$\bullet$] Nothing else is a conditional plan.
\end{itemize}
\end{definition}
\st Intuitively, to execute a plan $a; case(\varphi_1 \rightarrow
c_1, \ldots, \varphi_n \rightarrow c_n)$, first $a$ is executed.
$\varphi_i$'s are then evaluated. If one of $\varphi_i$ is true
then $c_i$ is executed. If none of $\varphi_i$ is true then the
plan fails. To execute a plan $c_1; c_2$, first $c_1$ is executed
then $c_2$ is executed. In Section \ref{sub1}, we formally define
the progression function $\Phi$ that encodes this intuition.

{\example [Getting to Evanston] \label{ex0}
The following is a conditional plan:\\
\st \hspace*{0.25in} $check\_traffic;\\
\hspace*{0.3in} case(\ \\
\hspace*{0.5in} traffic\_bad \rightarrow \\
\hspace*{0.7in}  goto\_western\_at\_belmont;\\
\hspace*{0.7in}  take\_belmont;\\
\hspace*{0.7in}  take\_ashland\\
\hspace*{0.5in} \neg traffic\_bad \rightarrow \\
\hspace*{0.7in}  goto\_western\_at\_belmont; \\
\hspace*{0.7in} take\_western\\
\hspace*{0.3in} ) $ \hfill $\Box$
}

\subsection{0-Approximation}
\label{sub1}

The 0-approximation in~\cite{SB01} is defined by a transition
function $\Phi$ that maps pairs of actions and approximate states
into sets of approximate states. We now present the necessary
notions and basic definitions of 0-approximation as follows.

\st {\bf Basic definitions and notations}:

\begin{itemize}
    \item [$\bullet$] A-state: An {\em approximate state} (or a-state) is a pair $\langle T,F
    \rangle$ where $T {\subseteq} A$ and $F {\subseteq} A$ are two
    disjoint sets of fluents.
    \item [$\bullet$] True, false, unknown:  Given an a-state $\sigma
    {=} \langle T,F \rangle$, $T$ (resp. $F$), denoted by $\sigma.T$
    (resp. $\sigma.F$), is the set of fluents which are true (resp.
    false) in $\sigma$; and $A {\setminus} (T \cup F)$ is the set
    of fluents which are unknown in $\sigma$. Given a fluent $f$, we
    say that $f$ is true (resp. false) in $\sigma$ if $f
    \in T$ (resp. $f \in F$). $f$ (resp. $\neg f$) holds in $\sigma$
    if $f$ is true (resp. false) in $\sigma$. $f$ is known (resp.
    unknown) in $\sigma$ if $f \in (T \cup F)$ (resp. $f \not \in (T
    \cup F)$). A set $L$ of fluent literals holds in an a-state
    $\sigma = \langle T,F \rangle$ if every member of $L$ holds in
    $\sigma$. A set $X$ of fluents is known in $\sigma$ if every
    fluent in $X$ is known in $\sigma$. An action $a$ is {\em
    executable} in $\sigma$ if $Pre_a$ holds in $\sigma$.
    \item [$\bullet$] Notations: Let $\sigma_1 {=} \langle T_1,F_1 \rangle$ and $\sigma_2 {=}
    \langle T_2,F_2 \rangle$ be two a-states.

    \begin{enumerate}
        \item $\sigma_1 {\cap} \sigma_2 {=} \langle T_1 {\cap} T_2, F_1 {\cap} F_2 \rangle$ is
        called the intersection of $\sigma_1$ and $\sigma_2$.
        \item We say $\sigma_1$ extends $\sigma_2$, denoted by $\sigma_2 {\preceq}
        \sigma_1$ if $T_2 {\subseteq} T_1$ and $F_2 {\subseteq} F_1$.
        $\sigma_1 {\setminus} \sigma_2$ denotes the set $(T_1 {\setminus}
        T_2) {\cup} (F_1 {\setminus} F_2)$.
        \item For a set of fluents $X$, we
        write $X {\setminus} \langle T,F \rangle$ to denote $X {\setminus}
        (T {\cup} F)$. To simplify the presentation, for a set of literals
        $L$, by $L^+$ and $L^-$ we denote the set of fluents $\{f \mid
        f{\in}L,\; f $ is a fluent $\}$ and  $\{f \mid \neg f{\in}L,\; f $
        is a fluent $\}$.
    \end{enumerate}
\end{itemize}


\st The transition function (for progression) is defined next.

\begin{definition} [Transition Function] \label{def-tran}
For an a-state $\sigma=\langle T,F \rangle$ and an action $a$,
$\Phi(a,\sigma)$ is defined as follows:
\begin{itemize} 
  \item [$\bullet$] if $a$ is not executable in $\sigma$ then $\Phi(a,\sigma) =
  \{\bot\}$; otherwise
  \item [$\bullet$] if $a$ is a non-sensing action:
    $\Phi(a,\sigma) =
\{\langle T \setminus Del_a \cup Add_a, F \setminus Add_a \cup
Del_a \rangle\};$
  \item [$\bullet$] if $a$ is a sensing action:
    $\Phi(a,\sigma) = \{\sigma' | \sigma \preceq \sigma'$ and $Sens_a \setminus \sigma = \sigma' \setminus \sigma\}$.
\end{itemize}
\end{definition}

\st The next example illustrates the above definition.

\begin{example}
[Getting to Evanston] \label{ex2} Consider an a-state
\[\sigma=\langle \{at\hbox{-}start\},\{on\hbox{-}western,
on\hbox{-}belmont,on\hbox{-}ashland, at\hbox{-}evanston \}
\rangle.\] We have that $check\_traffic$ is executable in $\sigma$
and
\[\Phi(check\_traffic, \sigma)= \{\sigma_1, \sigma_2\}\] where:\\
$\sigma_1 = \langle \{at\hbox{-}start, traffic\hbox{-}bad\},
\{on\hbox{-}western, on\hbox{-}belmont, on\hbox{-}ashland,
at\hbox{-}evanston \} \rangle$,\\ $\sigma_2 = \langle
\{at\hbox{-}start\},\{traffic\hbox{-}bad, on\hbox{-}western,
on\hbox{-}belmont, on\hbox{-}ashland, at\hbox{-}evanston\}
\rangle$.

\st Similarly, \[\Phi(goto\_western\_at\_belmont,\sigma) =
\{\sigma_3\}\] where: $\sigma_3 = \langle
\{$on{-}western,on{-}belmont$\}, \{$at{-}start, on{-}ashland,
at{-}evanston $\} \rangle$. \hfill $\Box$
\end{example}
\st The function $\Phi$ can be extended to define the function
$\Phi^*$ that maps each pair of a conditional plan $p$ and
a-states $\sigma$ into a set of a-states, denoted by
$\Phi^*(p,\sigma)$. $\Phi^*$ is defined similarly to
$\hat\Phi$ in~\cite{SB01}.

\begin{definition} [Extended Transition Function] \label{def-extran}
The extended transition function $\Phi^*$ is defined as follows:
\begin{itemize}
    \item [$\bullet$] For an empty sequence of actions and an a-state $\sigma$: $\Phi^*([\ ], \sigma) = \{\sigma\}$.
    \item [$\bullet$] For a non-sensing action $a$ and an a-state $\sigma$: $\Phi^*(a, \sigma) = \Phi(a,\sigma)$.
    \item [$\bullet$] For a case plan
          $c=a;case(\varphi_1 \rightarrow p_1, \ldots, \varphi_n \rightarrow p_n)$ where $a$ is a sensing action:
        \[\Phi^*(c, \sigma) = \bigcup_{\sigma' \in \Phi(a,\sigma)} E(case(\varphi_1 \rightarrow p_1, \ldots, \varphi_n \rightarrow p_n), \sigma')\] where
            \[ E(case(\varphi_1 \rightarrow p_1, \ldots, \varphi_n \rightarrow p_n), \gamma) =
            \left\{%
            \begin{array}{ll}
            \Phi^*(p_j,\gamma), & \hbox{if $\varphi_j$ holds in $\gamma$ ($1 \leq j \leq n$);} \\
            \{\bot\}, & \hbox{if none of $\varphi_1, \ldots, \varphi_n$ holds in $\gamma$.} \\
            \end{array}%
            \right.    \]
    \item [$\bullet$] For two conditional plans $c_1$ and $c_2$:
          $\Phi^*(c_1;c_2, \sigma) = \bigcup_{\sigma' \in \Phi^*(c_1,\sigma)} \Phi^*(c_2, \sigma')$.
    \item [$\bullet$] For any conditional plan $c$: $\Phi^*(c, \bot) = \{\bot\}$.
\end{itemize}
\end{definition}

\st Intuitively, $\Phi^*(c,\sigma)$ is the set of a-states
resulting from the execution of $c$ in $\sigma$.

\st Given a planning problem  $P=\langle A,O,I,G \rangle$, the
a-state representing $I$ is defined by $\sigma_I = \langle I^+,
I^- \rangle $. $\Sigma_G = \{\sigma \mid \sigma_G \preceq
\sigma\}$, where $\sigma_G = \langle G^+, G^- \rangle$, is the set
of a-states satisfying the goal $G$. A {\em progression solution}
to the planning problem $P$ is a conditional
plan $c$ such that 
$\Phi^*(c,\sigma_I) \subseteq \Sigma_G$. Note that, since $\bot$
is not a member of $\Sigma_G$, therefore $\bot \not\in
\Phi^*(c,\sigma_I)$.

{\example [Getting to Evanston - cont'd] \label{ex1}

\st Consider an initial state and goal states represented by the
sets:

\st \hspace*{0.25in} $I = \{at$-$start, \neg on$-$western, \neg
on$-$belmont, \neg on$-$ashland, \neg at$-$evanston\}$;

\st \hspace*{0.25in} $G = \{at$-$evanston\}$,

\st respectively. The following conditional plan (Example
\ref{ex0}) is a progression solution:

\st \hspace*{0.25in} $check\_traffic;\\
\hspace*{0.3in} case(\ \\
\hspace*{0.5in} traffic\_bad \rightarrow \\
\hspace*{0.7in}  goto\_western\_at\_belmont;\\
\hspace*{0.7in}  take\_belmont;\\
\hspace*{0.7in}  take\_ashland\\
\hspace*{0.5in} \neg traffic\_bad \rightarrow \\
\hspace*{0.7in}  goto\_western\_at\_belmont; \\
\hspace*{0.7in} take\_western\\
\hspace*{0.3in} ) $ \hfill $\Box$

}

\section{Regression and Its Relation with Progression} \label{Reg}

In this section, we will present our formalization of a regression
function, denoted by $Regress$, and prove that it is both sound
and complete with respect to the progression function $\Phi$.
$Regress$ is a state based regression function that maps a pair of
an action and a set of a-states into an a-state.

\st In our formulation, observe that given a plan $p$ and an
a-state $\sigma$, a goal $G$ is satisfied after the execution of
$p$ in $\sigma$ if $G$ holds in {\em all} a-states belonging to
$\Phi^*(p,\sigma)$, i.e., $G$ holds in $\cap_{\sigma' \in
\Phi^*(p,\sigma)} \sigma'$. This stipulates us to introduce the
notion of a {\em partial state} (or p-state) as a pair $[T,F]$
where $T$ and $F$ are two disjoint sets of fluents. 
Intuitively, a p-state $\delta {=}
[T,F]$ represents a collection of a-states which extends the
a-state $\langle T, F \rangle$. We denote this set by
$ext(\delta)$ and call it {\it the extension set} of $\delta$.
Formally, $ext(\delta) = \{\langle T', F' \rangle | T \subseteq
T', F \subseteq F'\}$. A $\sigma' \in ext(\delta)$ is called an
extension of $\delta$. Given a p-state $\delta {=} [T,F]$, we say
a partial state $\delta' {=} [T',F']$ is a partial extension of
$\delta$ if $T \subseteq T', F \subseteq F'$.


\st The regression function will be defined separately for
non-sensing actions and sensing actions. Since the application of
a non-sensing action in an a-state results into a single a-state,
the regression of a non-sensing action should be with respect to a
p-state and result in a p-state. On the other hand, since the
application of a sensing action in an a-state results in a set of
a-states, the regression of a sensing action should be with
respect to a set of p-states and result in a p-state. Besides the
regression should be sound (i.e., plans obtained through
regression must be plans based on the progression) and complete
(i.e., for each plan based on progression, using regression one
should obtain that plan or an equivalent one) with respect to
progression. We will now formulate this notion precisely.

\st We adopt the use of the term ``application'' \cite{BG01} in
formulating regression to distinguish from the use of
``execution'' in progression. To simplify the presentation, we
define a partition of a set of fluents $X$ as a pair $(P,Q)$ such
that $P \cap Q = \emptyset$ and $P \cup Q = X$. We begin with the
applicability condition of non-sensing actions and then give the
definition of the function $Regress$ for non-sensing actions.
\begin{definition}
[Applicability Condition - non-sensing action] Given a non-sensing
action $a$ and a p-state $\delta=[ T,F ]$. We say that $a$ is
    {\em applicable} in $\delta$ if (i) $Add_a \cap T \neq \emptyset$
    or $Del_a \cap F \neq \emptyset$, and (ii)
    $Add_a \cap F = \emptyset$, $Del_a \cap T = \emptyset$,
    $Pre^+_a \cap F \subseteq Del_a$, and $Pre^-_a \cap T \subseteq Add_a$.
\end{definition}
Intuitively, the applicability condition aforementioned is a
\emph{relevance} (item (i)) and \emph{consistency} condition (item
(ii)) for $a$. Item (i) is considered ``relevant'' as it makes
sure that the effects of $a$ will contribute to $\delta$ after
execution. Item (ii) is considered ``consistent'' as it makes sure
that the situation obtained by progressing $a$, from a situation
yielded by regressing $a$ from $\delta$, will be consistent with
$\delta$.

\st The regression on a non-sensing action is defined next.
\begin{definition} [Regression - non-sensing action] \label{reg-nonsensing} Given
a non-sensing action $a$ and a p-state $\delta = [T,F]$,
\begin{list}{$\bullet$}{\topsep=2pt \itemsep=2pt \parsep=2pt}
\item if $a$ is not applicable in
    $\delta$ then $Regress(a,\delta) = \bot$;
\item if $a$ is applicable in $\delta$ then {$ Regress(a,\delta) =
[ T \setminus Add_a \cup Pre_a^+, F \setminus Del_a \cup Pre_a^-
]$.}
\end{list}
\end{definition}
\ni For later use, we extend the regression function $Regress$ for
non-sensing actions over a set of p-states and define
 $$Regress(a,\{\delta_1, \ldots, \delta_n\}) =
\{Regress(a,\delta_1), \ldots, Regress(a,\delta_n)\}$$

\st where $\delta_1, \ldots, \delta_n$ are p-states and $a$ is a
non-sensing action.

{\example [Getting to Evanston -  con't] \label{ex4}

The actions $take\_western$ and $take\_ashland$ are applicable in
$\delta = [\{$at{-}evanston$\},\{\}]$.

\st \hspace*{0.25in} $Regress(take\_western, \delta) =
[\{$on{-}western$\}, \{$traffic{-}bad $\}]$, and

\st \hspace*{0.25in} $Regress(take\_ashland, \delta) =
[\{$on{-}ashland$\}, \{\}]$. \hfill $\Box$}

\ni We will now define $Regress$ for sensing actions. Recall that
the execution of a sensing action $a$ in an a-state $\sigma$
requires that $a$ is executable in $\sigma$ and results in a set
of a-states $\Phi(a,\sigma)$ whose member extends $\sigma$ by the
set of fluents in $s_a \subseteq Sens_a$ and every $f \in Sens_a
\setminus s_a$ is known in $\sigma$. This leads to the following
definitions.

\begin{definition} [Properness] \label{sensed-Set} Let $a$ be a sensing action,
$\Delta= \{\delta_1, \ldots, \delta_n\}$ be a set of distinct
p-states, and $\emptyset \neq X \subseteq Sens_a$ be a set of
sensing fluents. We say that $\Delta$ is proper with respect to
$X$ if (i) $Sens_a$ is known in $\Delta$; (ii) $n= 2^{|X|}$; (iii)
for every partition $(P,Q)$ of $X$, there exists only one
$\delta_i \in \Delta$ ($1 \leq i \leq n$) such that. $\delta_i.T
\cap X = P,\ \delta_i.F \cap X = Q$; and (iv) for every ($1 \leq i
\neq j \leq n$), $\delta_i.T \setminus X = \delta_j.T \setminus
X$, $\delta_i.F \setminus X = \delta_j.F \setminus X$. We call $X$
as a sensed set of $\Delta$ with respect to $a$.

\end{definition}

{\example [Getting to Evanston -  con't] \label{ex41} Consider a
set $\Delta_1 = \{ \delta_1, \delta_2\}$ where $\delta_1 = [
\{$at{-}start, traffic{-}bad$\}, \{$on{-}western, on{-}belmont,
on{-}ashland, at{-}evanston$\} ]$ and $\delta_2 = [
\{$at{-}start$\}, \{$traffic{-}bad, on{-}western, on{-}belmont,
on{-}ashland, at{-}evanston$\} ]$.

\st We have that $\Delta_1$ is proper with respect to
$\{traffic$-$bad\}$. The set $\{traffic$-$bad\}$ is the sensed set
of $\Delta_1$ with respect to $check$-$traffic$.

\st Consider $\Delta_2 = \{\delta_1, \delta_3\}$ where $\delta_3 =
[ \{$at{-}start$\}, \{$traffic{-}bad, at{-}evanston$\} ]$. We have
that $\Delta_2$ is not proper with respect to
$\{traffic$-$bad\}$. \hfill $\Box$}

{\lemma [Sensed Set] \label{lemSS} Consider a sensing action $a$
and a set of p-states $\Delta$. If there exists a sensed set of
$\Delta$ with respect to $a$ then it is unique.}

\st {\textbf{Proof:} In Appendix.}

\st Given a sensing action $a$ and a set of p-states $\Delta$, we
denote $p(a,\Delta)$ as the unique sensed set of $\Delta$ with
respect to $a$; if there exists no sensed set with respect to $a$
and $\Delta$, we write $p(a,\Delta) = \bot$.

\begin{definition} [Strong Applicability Condition - sensing action]
\label{app-sensing-strong}  Let $a$ be a sensing action and
$\Delta=\{\delta_1, \ldots, \delta_n\}$ be a set of p-states. We
say that $a$ is \emph{strongly applicable} in $\Delta$ if (i)
$p(a,\Delta) \neq \bot$; and (ii) $Pre^+_a \cap \delta_i.F =
\emptyset$ and $Pre^-_a \cap \delta_i.T = \emptyset$.
\end{definition}
In the above definition, (i) corresponds to the fact that
executing a sensing action $a$ in an a-state $\sigma$ results in a
set of $2^{|p(a,\Delta)|}$ a-states that are represented by
$2^{|p(a,\Delta)|}$ corresponding p-states of $\Delta$ where
$p(a,\Delta)$ denotes the set of fluents that are not yet known,
while $Sens_a \setminus p(a,\Delta)$ is already known when $a$ is
executed; (ii) guarantees that $a$ must be executable prior to its
execution.

\st Although this strong applicability condition guarantees the
soundness of regression over sensing actions, it does not
guarantee the completeness. We now provide a weaker applicability
condition that guarantees both soundness and completeness of
regression.

\begin{definition} \label{app-sensing} [Applicability Condition - sensing action]
Let $a$ be a sensing action and $\Delta=\{\delta_1, \ldots,
\delta_n\}$ be a set of p-states. We say that $a$ is
\emph{applicable} in $\Delta$ if (i) there exists a set $\Delta'
{=} \{\delta'_1, \ldots, \delta'_n\}$, where $\delta'_i$ is a
partial extension of $\delta_i$ ($i=1, \ldots,n$), such that $a$
is strongly applicable in $\Delta'$; and (ii) $Sens_a$ is known in
$\Delta$.
\end{definition}

{\lemma [Unique Sensed Set] \label{lemSS1} Consider a sensing
action $a$ and a set of p-states $\Delta$ such that $a$ is
applicable in $\Delta$. Let $\Delta' {=} \{\delta'_1, \ldots,
\delta'_n\}$, where $\delta'_i$ is a partial extension of
$\delta_i$ ($i=1, \ldots,n$), $\Delta'' {=} \{{\delta_1}'',
\ldots, {\delta_n}''\}$, where ${\delta_i}''$ is a partial
extension of $\delta_i$ ($i=1, \ldots,n$). If $p(a,\Delta') \neq
\bot$ and $p(a,\Delta'') \neq \bot$ then $p(a,\Delta') =
p(a,\Delta'')$.}

\st {\textbf{Proof:} In Appendix.}

\st Given a sensing action $a$ and a set of p-states $\Delta$. If
there exists a $\Delta' {=} \{\delta'_1, \ldots, \delta'_n\}$,
where $\delta'_i$ is a partial extension of $\delta_i$ ($i=1,
\ldots,n$) such that $p(a,\Delta') \neq \bot$ then, by Lemma
\ref{lemSS1}, $p(a,\Delta') = p(a,\Delta'')$ for all $\Delta''=
\{{\delta_1}'', \ldots, {\delta_n}''\}$, where ${\delta_i}''$ is a
partial extension of $\delta_i$ ($i=1, \ldots,n$) and
$p(a,\Delta'') \neq \bot$. We refer to the set $p(a,\Delta')$ by
$S_{a,\Delta}$. If there exists no such $p(a,\Delta')$, we write
$S_{a,\Delta} = \bot$. Note that, from Definition
\ref{app-sensing}, if $a$ is applicable in $\Delta$ then
$S_{a,\Delta}$ is defined. In that case, we also often say that
$a$ is applicable in $\Delta$ with respect to $S_{a,\Delta}$ to
make the applicability condition clearer from the context.

{\example [Getting to Evanston -  con't] \label{ex42} Consider the
set $\Delta_2$ and the sensing action $check$-$traffic$ in Example
\ref{ex41}. We have that

\begin{itemize}
    \item [(i)] $check$-$traffic$ is not strongly applicable in $\Delta_2$ (w.r.t $traffic$-$bad$), however,
    \item [(ii)] $check$-$traffic$ is
applicable in $\Delta_2$ (w.r.t $traffic$-$bad$).
\end{itemize}
Note that $\Delta_1$ in Example \ref{ex41} consists of partial
extensions of p-states in $\Delta_2$, and $check$-$traffic$ is
strongly applicable in $\Delta_1$ (with respect to
$traffic$-$bad$). \hfill $\Box$}

\begin{definition} [Regression - sensing action] \label{reg-sensing}
Let $a$ be a sensing action and $\Delta=\{\delta_1, \ldots,
\delta_n\}$ be a set of p-states.
\begin{itemize} 
  \item [$\bullet$] if $a$ is not applicable in
    $\Delta$ then $Regress(a, \Delta) = \bot$; and
  \item [$\bullet$] if $a$ is applicable in
    $\Delta$ \\
\centerline{ $ Regress(a, \Delta) =
    [ (\bigcup^n_{i=1}\delta_i.T) \setminus S_{a,\Delta} \cup Pre^+_a, (\bigcup^n_{i=1}\delta_i.F) \setminus S_{a,\Delta} \cup Pre^-_a ]$.}
\end{itemize}
\end{definition}

{\example [Getting to Evanston -  con't] \label{ex5} \st
$check\_traffic$ is applicable in $\Delta_2$ with respect to
$\{traffic\_bad\}$ (see Example \ref{ex42}) and we have

\hspace{.25in} $Regress(check\_traffic, \Delta_2) =$

\hspace{.65in} $[\{at\hbox{-}start\},\{on\hbox{-}western,
on\hbox{-}belmont, on\hbox{-}ashland, at\hbox{-}evanston\} ].$\hfill $\Box$}

\st We now relate our regression function $Regress$ with the
progression function $\Phi$.

\subsection{Soundness Result}

\begin{figure}[h]
    \centerline{
    \begin{tabular}{c}
    {
    \epsfxsize=3.5in \epsfbox{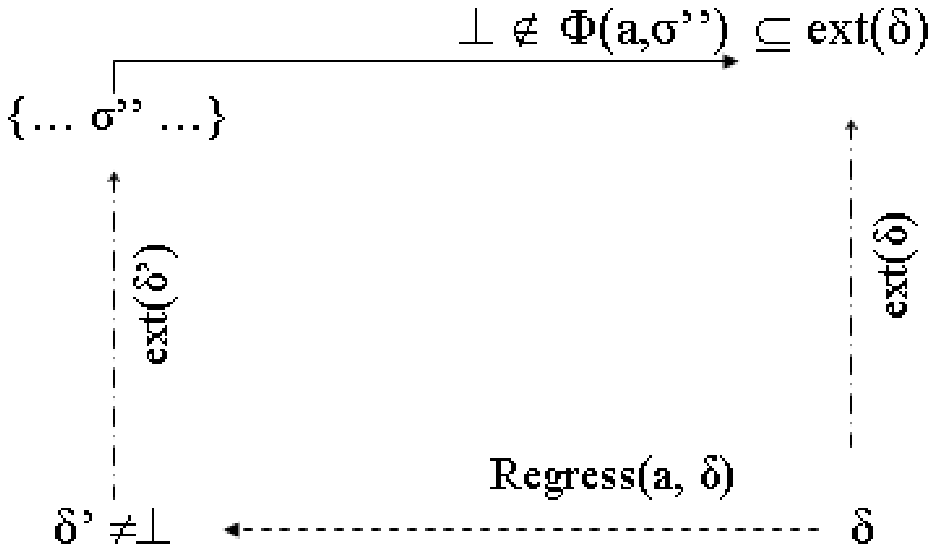}
    }
    \end{tabular}}
    \caption{Illustration of Proposition \ref{lem1}.}
    \label{example-fig}
\end{figure}

{\proposition [Non-sensing action] \label{lem1} Let $\delta$ and
$\delta'$ be two p-states, and $a$ be a non-sensing action. If
$Regress(a,\delta)= \delta'$ and $\delta' \neq \bot$, then for
every $\sigma'' \in ext(\delta')$ we have that (i) $\bot \not \in
\Phi(a, \sigma'')$, and (ii) $\Phi(a, \sigma'') \subseteq
ext(\delta)$.}

\st {\textbf{Proof:} In Appendix.}

\st Intuitively, this proposition states that the regression of a
non-sensing action in a p-state yields another p-state such that
the execution of the action in any extension of the latter results
in a subset of a-states belonging to the extension set of the
former. This shows that $Regress$ can be ``reversed'' for
non-sensing actions.

{\proposition [Sensing action] \label{lem2} Let
$\Delta=\{\delta_1, \ldots, \delta_n\}$ be a set of p-states,
$\delta'$ be a p-state, and $a$ be a sensing action. If
$Regress(a,\Delta) = \delta'$ where $\delta' \neq \bot$, then for
every $\sigma'' \in ext(\delta')$, we have that (i) $\bot \not \in
\Phi(a, \sigma'')$, and (ii) $\Phi(a,\sigma'') \subseteq
ext(\delta_1) \cup \ldots \cup ext(\delta_n)$.}

\st {\textbf{Proof:} In Appendix.}

\st Similarly, this proposition states that the regression of a
sensing action in a set of p-states yields a p-state such that the
execution of the action in any extension of the latter results in
a subset of a-states belonging to the union of the extension sets
of the formers. This also shows that $Regress$ can be ``reversed''
for sensing actions.

\begin{figure}[h]
    \centerline{
    \begin{tabular}{c}
    {
    \epsfxsize=3.9in \epsfbox{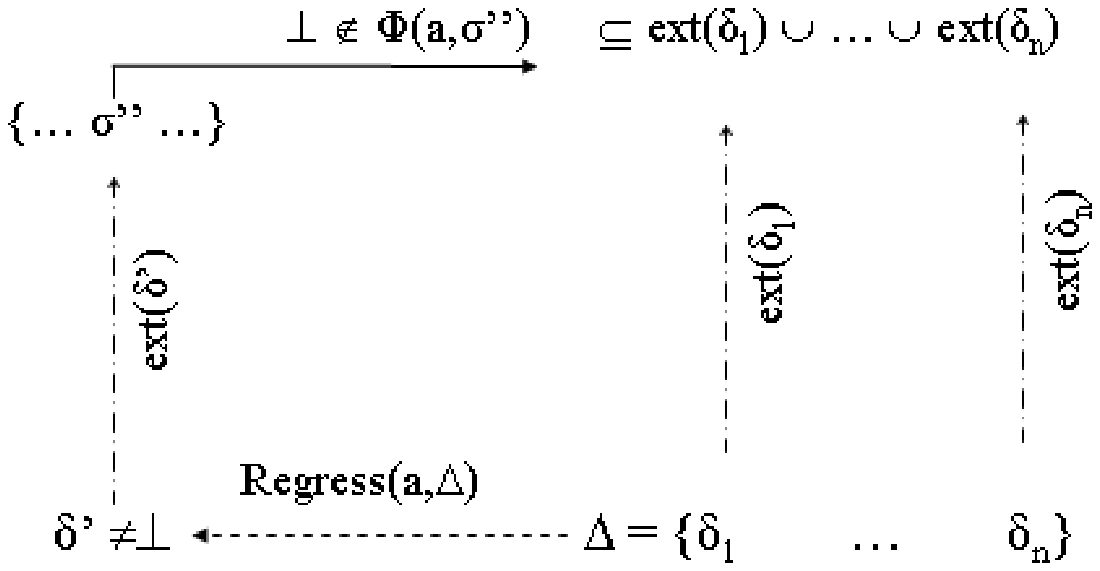}
    }
    \end{tabular}}
    \caption{Illustration of Proposition \ref{lem2}.}
    \label{example-fig}
\end{figure}

\st We next extend $Regress$ to define $Regress^*$ that allows us
to perform regression with respect to conditional plans.

\begin{definition}[Extended Regression Function] \label{d3}
Let $\delta$ and $\{\delta_1, \ldots, \delta_n\}$ be a p-state and
a set of p-states, respectively. The extended transition function
$Regress^*$ is defined as follows:

\begin{itemize} 

\item [$\bullet$] $Regress^*([\ ], \delta) = \delta$.

\item [$\bullet$] For a non-sensing action $a$, $Regress^*(a,
\delta)$ = $Regress(a, \delta)$.

\item [$\bullet$] For a conditional plan $p  = a;case(\varphi_1
{\rightarrow} c_1, \ldots, \varphi_n {\rightarrow} c_n)$,

\begin{itemize} 

\item [--] if $Regress^*(c_i, \delta) {=} \bot$ for some $i$, $
Regress^*(p,\delta) = \bot$;

\item [--] if $Regress^*(c_i, \delta) {=} [T_i,F_i]$
$i=1,\ldots,n$, then \\
\centerline{$ Regress^*(p,\delta) = Regress(a,\{ R(c_1,\delta),
\ldots, R(c_n,\delta)\})$} where $R(c_i, \delta) = [T_i \cup
\varphi_i^+,F_i \cup \varphi_i^-]$ if $\varphi_i^+ \cap F_i =
\emptyset$ and $\varphi_i^- \cap T_i = \emptyset$; otherwise,
$R(c_i, \delta) = \bot$. Here, $\varphi_i^+$ and $\varphi_i^-$
denote the sets of fluents occurring positively and negatively in
$\varphi_i$, respectively.
\end{itemize}

\item [$\bullet$] For $p = c_1; c_2$, where $c_1, c_2$ are conditional plans, \\
\centerline{ $Regress^*(p,\delta) = Regress^* (c_1,Regress^* (c_2,
\delta))$;}
\item [$\bullet$] $Regress^* (p,\perp) = \perp$ for
every plan $p$.
\end{itemize}
\end{definition}

\st For a planning problem $P=\langle A,O,I,G \rangle$, let
$\delta_G$ be the p-state $[G^+, G^- ]$, and $\Delta_I$ is the set
of p-states such that for every $\delta \in \Delta_I$, $\sigma_I
\in ext(\delta)$. (Recall that $\sigma_I$ is the a-state
representing $I$ and $\Sigma_G$ is the set of a-states in which
$G$ holds). A {\it regression solution} to the planning problem
$P$ is a conditional plan $c$ that upon applying from the p-state
$\delta_G$ will result in one of the p-states in $\Delta_I$. In
other words, if $\delta = Regress^*(c, \delta_G)$ then $\delta$ is
a p-state belonging to $\Delta_I$.

\st We now formalize the following relationship between the
regression function $Regress^*$ with the progression transition
function $\Phi^*$.

{\theorem [Soundness of Regression] \label{lem3} Let  $P=\langle
A,O,I,G \rangle$ be a planning problem and $c$ be a regression
solution of $P$. Then, $c$ is also a progression solution of $P$,
i.e., $\bot \not\in \Phi^*(c,\sigma_I)$ and $\Phi^*(c,\sigma_I)
\subseteq ext(\delta_G)$. }

\st {\textbf{Proof:} In Appendix.}

\begin{figure}[h]
    \centerline{
    \begin{tabular}{c}
    {
    \epsfxsize=3.7in \epsfbox{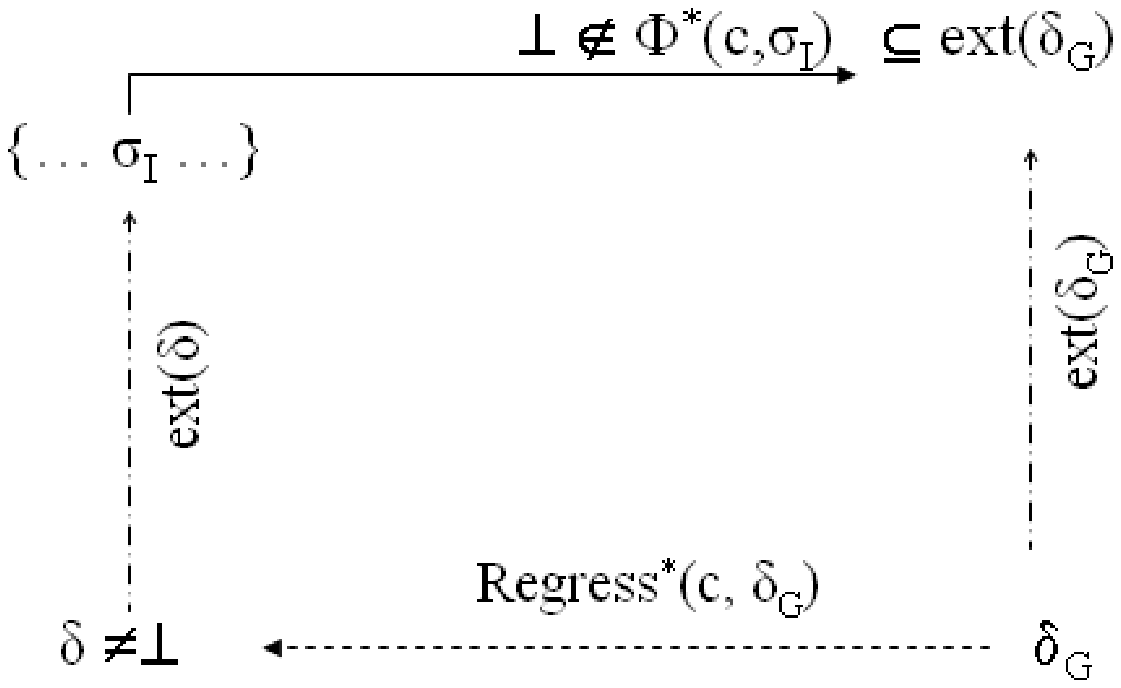}
    }
    \end{tabular}}
    \caption{Illustration of Theorem \ref{lem3}.}
    \label{example-fig}
\end{figure}

\subsection{Completeness Result}

\st We now proceed towards a completeness result. Ideally, one
would like to have a completeness result that expresses that for a
given planning problem, any solution found through progression can
also be found by regression. In our formulation, however, the
definition of the progression function allows an action $a$ to
execute in any a-state $\sigma$ if $a$ is executable in $\sigma$,
regardless whether $a$ would add ``new'' information to $\sigma$
or not. In contrary, our definition of the regression function
requires that an action $a$ can only be applied in a state (or a
set of states) if $a$ contributes effects to the applied state(s)
\footnote{Note that this condition is also applied for regression
planning systems such as \cite{BG01}, \cite{nguyenetal02}.}. Thus,
given a planning problem $P = \langle A,O,I,G \rangle$, a
progression solution $c$ of $P$ may contains redundant actions or
extra branches. As a result, we may not obtain $c$ via our
regression, i.e. $Regress^*(c, \delta_G) = \bot$. To illustrate
the aforementioned points, let's consider the following two
examples. Example \ref{EXredundancy-action} shows conditional
plans, each is a sequence of non-sensing actions, that contain
redundant actions.

{\example [Redundancy] \label{EXredundancy-action} Let $P =
\langle \{f,g\}, \{a,b\}, \{f\}, \{g\} \rangle$ be a planning
problem where $a$ is a non-sensing action with $Pre_a = \{f\}$,
$Add_a = \{g\}$, and $Del_a = \emptyset$; $b$ is also a
non-sensing action where $Pre_b = \{g\}$, $Add_b = \{f\}$, and
$Del_b = \emptyset$. Clearly $a$, $a;b$, and $a;a$ are progression
solutions achieving the goal $\{g\}$. However, we can see that
$b$, and a copy (a.k.a an instance) of $a$ in the second and third
plans, respectively, are redundant.

\st We also have that $Regress^*(a;b, [\{g\}, \emptyset]) = \bot$
and $Regress^*(a;a, [\{g\}, \emptyset]) = \bot$. Note that
$Regress^*(a, [\{g\}, \emptyset]) \neq \bot$. \hfill $\Box$}

\st Example \ref{EXredundancy-action2} shows a conditional plan
that contains redundant branches in a case plan.

{\example [Redundancy] \label{EXredundancy-action2} Let $P' =
\langle \{f',g'\}, \{a',b'\}, \{f'\}, \{g'\} \rangle$ be a
planning problem. Let $a'$ be a sensing action where $Pre_{a'} =
\emptyset$, $Sens_{a'} = \{f',g'\}$; $b'$ is a non-sensing action
where $Pre_{b'} = \{f', \neg g'\}$, $Add_{b'} = \{g'\}$, and
$Del_{b'} = \emptyset$. A plan achieving $g'$ is:
\[p' = a'; case(f' \wedge \neg g' \rightarrow b', f' \wedge g'
\rightarrow [\ ], \neg f' \wedge \neg g' \rightarrow [\ ], \neg f'
\wedge g' \rightarrow [\ ]).\] Notice that, the two branches with
conditions $\neg f' \wedge \neg g'$ and $\neg f' \wedge g'$ that
are always evaluated to false when $a$ get executed, and thus are
never used to achieve $g'$.

\st We also have that $Regress^*(p', [\{g'\}, \emptyset]) = \bot$.
Let $p''$ be the conditional plan obtained from $p'$ by removing
two branches with conditions $\neg f' \wedge \neg g'$ and $\neg f'
\wedge g'$:

\[p'' = a'; case(f' \wedge \neg g' \rightarrow b', f' \wedge g'
\rightarrow [\ ]).\]

\st Then, $Regress^*(p'', [\{g\}, \emptyset]) \neq \bot$. \hfill
$\Box$}

\st The above discussion stipulates us to consider the following
completeness result: if a conditional plan can be found through
progression we can find an equivalent conditional plan through
regression. The plan found through regression does not have
redundancies, both in terms of extra actions and extra branches.
We refer to these notions as ``redundancy'' and ``plan
equivalence''. We now formalize these notions. First we need the
following notion. Given a sensing action $a$,  a sub sensing
action of $a$ is a sensing action $a'$ where $Pre_{a'} = Pre_a$
and $Sens_{a'} \subset Sens_a$.


\begin{definition}[Subplan] \label{subplan}
Let $c$ be a conditional plan. A conditional plan $c'$ is a
subplan of $c$ if
\begin{itemize} 
    \item [$\bullet$] $c'$ can be obtained from $c$ by
    ({\em i}) removing an instance of a non-sensing action from
    $c$; or ({\em ii}) removing a case plan or a branch
    $\varphi_i \rightarrow c_i$ from a case plan in $c$; or
    ({\em iii}) replacing a sensing action $a$ with a sub
    sensing action $subSense(a)$ of $a$; or
    \item [$\bullet$] $c'$ is a subplan of $c''$ where $c''$
    is a subplan of $c$.
\end{itemize}
\end{definition}

\begin{definition}[Redundancy] \label{redundancy} Let $c$ be a conditional
plan, $\sigma$ be an a-state, and $\delta$ be a p-state. We say
that $c$ contains redundancy (or is redundant) with respect to
$(\sigma,\delta)$ if
\begin{itemize}
    \item [(i)] $\bot \not \in \Phi^*(c,\sigma)$ and
        $\Phi^*(c,\sigma) \subseteq ext(\delta)$; and
    \item [(ii)] there exists a subplan $c'$ of $c$ such that
          $\bot \not \in \Phi^*(c',\sigma)$ and $\Phi^*(c',\sigma) \subseteq ext(\delta)$.
\end{itemize}
\end{definition}

\st Note that, if $c'$ is a subplan of a conditional plan $c$ then
$c' \neq c$. The equivalence of two conditional plans is defined
formally as follows.

\begin{definition} [Equivalent Plan] \label{equivPlan}
Let $\sigma$ be an a-state, $\delta$ be a p-state, and $c$ be a
conditional plan such that $\bot \not \in \Phi^*(c,\sigma)$ and
$\Phi^*(c,\sigma) \subseteq ext(\delta)$. We say a conditional
plan $c'$ is equivalent to $c$ with respect to $(\sigma,\delta)$
if $\bot \not \in \Phi^*(c',\sigma)$ and $\Phi^*(c',\sigma)
\subseteq ext(\delta)$.
\end{definition}

{\example [Equivalence] \label{EXredundancy-action3} Consider the
planning problem $P$ in Example \ref{EXredundancy-action}, we have
that $a$ is a subplan of $a;a$ and is equivalent to $a;a$ w.r.t
$(\langle \{f\},\emptyset \rangle, [\{g\}, \emptyset])$.

\st Similarly, for planning problem $P'$ (Example
\ref{EXredundancy-action2}), $p''$ is a subplan of $p'$ and is
equivalent to $p'$ w.r.t $(\langle \{f'\},\emptyset \rangle,
[\{g'\}, \emptyset])$. \hfill $\Box$}

\st It's easy to see that if there exist conditional plans $c'$
and $c''$ that are both equivalent to $c$ with respect to
$(\sigma,\delta)$ then $c'$ and $c''$ are equivalent with respect
to $(\sigma,\delta)$. 

\st We will continue with our formulation. Recall that our purpose
is to use regression to find an equivalent conditional plan for a
given progression solution. To do that, we will introduce a notion
called \emph{normalized conditional plans}. Such conditional plans
can be generated by our planning algorithm which is introduced in
Section \ref{algo}. We will also need to provide conditions about
when a conditional plan is regressable, i.e. when $Regression^*$
function can be applied on it to produce a p-state. We refer to
conditional plans with such conditions as \emph{regressable
conditional plans}. We will later show that, for a given
progression solution of a planning problem $P$ there always exists
an equivalent normalized, regressable conditional plan that is
also a regression solution of $P$. The definition of a normalized
conditional plan is as follows.

\begin{definition} [Normalized Conditional Plan] \label{normalizedPlan}
A conditional plan $c$ is a \emph{normalized conditional plan} if
$c = \alpha ; c'$ where $\alpha$ is the empty plan or a sequence
of non-sensing actions, $c' = [\ ]$ or $c' = a; case(\varphi_1
\rightarrow p_1, \ldots, \varphi_m \rightarrow p_m)$, $a$ is a
sensing action, and the $p_i$'s are normalized conditional plans.
\end{definition}

{\example [Normalized Plan] \label{EXnormalized} Consider the
planning problem $P'$ in Example \ref{EXredundancy-action2}. Then
both $p'$ and $p''$ is a normalized conditional plan. Note that
$p''$ is regressable whilst $p'$ is not regressable. \hfill
$\Box$}

\st For a plan $c = c_1;\ldots;c_n$ where $c_i$ is either a
sequence of actions or a case plan, we define $normalized(c)$, a
normalized conditional plan obtained from $c$, as follows.

\begin{itemize}
\item [$\bullet$] If $n = 1$ and $c_1$ is a sequence of
non-sensing actions then $normalized(c) = c$.

\item [$\bullet$] If $n = 1$ and $c_1$ is a case plan, $c_1 = a;
case\ (\varphi_1 \rightarrow p_1
        \ldots \varphi_m \rightarrow p_m\ )$,
then $normalized(c) = a; case\ (\varphi_1 \rightarrow
normalized(p_1)
        \ldots \varphi_m \rightarrow normalized(p_m)\ )$.

\item [$\bullet$] If $n > 1$ and $c_1$ is a sequence of
non-sensing actions then $normalized(c) = c_1; normalized(c')$
where $c' = c_2;\ldots;c_n$.

\item [$\bullet$] If $n > 1$ and $c_1$ is a case plan, $c_1 = a;
case\ (\varphi_1 \rightarrow p_1
        \ldots \varphi_m \rightarrow p_m\ )$,
then $normalized(c) = a; case\ (\varphi_1 \rightarrow
normalized(p_1;c')
        \ldots \varphi_m \rightarrow normalized(p_m;c')\ )$
where $c' = c_2;\ldots;c_n$.
\end{itemize}

\ni The next lemma shows that for every conditional plan $c$ there
is an equivalent normalized conditional plan which is constructed
by the method above.

\begin{lemma} \label{lemAdd1}
For every conditional plan $c$,

\begin{itemize}
\item [$\bullet$] $normalized(c)$ is a normalized conditional
plan;

\item [$\bullet$] for every a-state $\sigma$, $\Phi^*(c,\sigma) =
\Phi^*(normalized(c), \sigma)$.

\end{itemize}
\end{lemma}

\st {\textbf{Proof:} In Appendix.}

\st To define a regressable conditional plan, we begin with some
additional notations. For a non-empty set of fluents  $S =
\{f_1,...,f_k\}$, a binary representation of $S$ is a formula of
the form $l_1 \wedge \ldots \wedge l_k$ where $l_i \in \{f_i, \neg
f_i\}$ for $i= 1, \ldots, k$.

\st For a non-empty set of fluents $S$, let $BIN(S)$ denote
the set of all different binary representations of $S$. We say a
conjunction $\phi$ of literals is consistent if there exists no
fluent $f$ such that $f$ and $\neg f$ appear in $\phi$. A set of
consistent conjunctions of literals $\chi = \{\varphi_1, \ldots,
\varphi_n\}$ is said to span over $S$ if there exists a consistent
conjunction of literals $\varphi \not \in \chi$, such that:

\begin{enumerate}
    \item $S \cap (\varphi^+ \cup \varphi^-) = \emptyset$ where
    $\varphi^+$ and $\varphi^-$ denote the sets of fluents occurring positive and negative in
    $\varphi$, respectively;
    \item $\varphi_i = \varphi \wedge \psi_i$ where $BIN(S) = \{ \psi_1, \ldots, \psi_n
    \}$.
\end{enumerate}

\st Notice that given a non-empty set $S$, we can easily check
whether the set $\chi = \{\varphi_1, \ldots, \varphi_n\}$ spans
over S. We say that a set $\chi = \{\varphi_1, \ldots,
\varphi_n\}$ is factorable if it spans over some non-empty set of
fluents $S$.

{\example [Getting to Evanston -  con't] \label{ex5a} Consider a
set $S= \{traffic$-$bad\}$, a conjunction $\varphi = on$-$ashland$
and a set of literal conjunctions $\chi = \{on$-$ashland \wedge
traffic$-$bad, on$-$ashland \wedge \neg traffic$-$bad\}$.

\st We have that $BIN(S) = \{traffic$-$bad, \neg traffic$-$bad\}$
and $\chi$ spans over $S$. \hfill $\Box$}

{\lemma \label{factorable} Let $\chi = \{\varphi_1, \ldots,
\varphi_n\}$ be a non-empty set of consistent conjunctions of
literals. If $\chi$ is factorable, then there exists a unique
non-empty set of fluents $S$ such that $\chi$ spans over $S$.}

\st {\textbf{Proof:} In Appendix.}

\begin{definition} [Possibly Regressable Case Plan] \label{Reg-CaseStructure} Given a
case plan $p = a; case(\varphi_1 \rightarrow c_1, \ldots,
\varphi_n \rightarrow c_n)$. We say that $p$ is possibly
regressable if (i) there exists a non-empty set $\emptyset \ne S_a
\subseteq Sens_a$ and $\{\varphi_1, \ldots, \varphi_n\}$ spans
over $S_a$, and (ii) for $1 \leq i \leq n$
    $Sens_a \subseteq (\varphi^+_i \cup \varphi^-_i)$.
\end{definition}

\begin{definition}[Regressable Conditional Plan] \label{RegConditionalPlan}
Let $c$ be a conditional plan, $\sigma$ be an a-state, and
$\delta$ be a p-state. We say $c$ is regressable with respect to
$(\sigma,\delta)$ if (i) every case plan occurring in $c$ is
possibly regressable, and (ii) $\bot \not \in \Phi^*(c,\sigma)
\subseteq ext(\delta)$ and $c$ is not redundant with respect to
$(\sigma,\delta)$.
\end{definition}

{ \lemma \label{lem31b} Let $\sigma$ be an a-state, $\delta$ be a
p-state, and $c$ is a normalized conditional plan that is
regressable with respect to $(\sigma,\delta)$. Then,
$Regress^*(c,\delta) = \delta'$, $\delta' \neq \bot$, and $\sigma
\in ext(\delta')$. }

\st {\textbf{Proof:} In Appendix.}

\st The following lemma shows conditions for the existence of a
normalized, regressable conditional plan that is equivalent to a
given normalized conditional plan.

\st {\lemma \label{lemAdd3} Let $\sigma$ be an a-state, let
$\delta$ be a p-state, and let $c$ be a normalized conditional
plan such that $\bot \not \in \Phi^*(c,\sigma)$ and
$\Phi^*(c,\sigma) \subseteq ext(\delta)$. There exists a
normalized plan $c'$ such that $c'$ is regressable with respect to
$(\sigma,\delta)$ and $c'$ is equivalent to $c$ with respect to
$(\sigma,\delta)$.}

\st {\textbf{Proof:} In Appendix.}

\st It follows from Lemma \ref{lemAdd1} and Lemma \ref{lemAdd3}
that, for every conditional plan $c$ there exists a normalized,
regressable conditional plan that is equivalent to $c$ under the
conditions mentioned in Lemma \ref{lemAdd3}. This provides a solid
building block for our completeness result. This result is
formally stated in the following theorem.

\begin{theorem} [Completeness of Regression] \label{theoAdd1}
Given a planning problem $P= \langle A,O,I,G \rangle$ and a
progression solution $c$ of $P$. There exists a normalized
regression solution $c'$ of $P$ such that $c'$ is equivalent to
$c$ with respect to $(\sigma_I, \delta_G)$.
\end{theorem}

\st {\textbf{Proof:} In Appendix.}

\begin{figure}[h]
    \centerline{
    \begin{tabular}{c}
    {
    \epsfxsize=3.2in \epsfbox{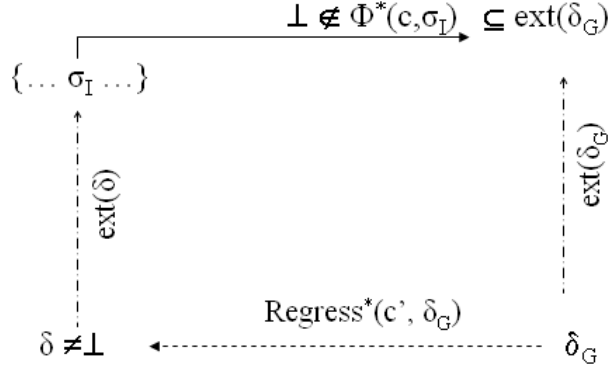}
    }
    \end{tabular}}
    \caption{Illustration of Theorem \ref{theoAdd1}.}
    \label{example-fig}
\end{figure}

\st We now present an algorithm that uses our regression functions
to construct conditional plans with sensing actions.

\section{Conditional Planning Using Regression} \label{algo}

In this section, we present a regression search algorithm for
constructing conditional plans with sensing actions that makes use
of the $Regress$ function described in the previous section. This
algorithm, while doing the search, records the plans used to get
to a p-state. For a conditional plan $c$ and a p-state $\delta$,
we call the pair $\langle c, \delta \rangle$ a {\em plan-state}
pair. For a set of plan-state pairs $X$, by $X_s$ we denote the
set of all the p-states occurring in $X$. The main idea of the
algorithm is as follows. At any step, we will maintain a set $N$
of plan-state pairs $\langle c, \delta \rangle$ such that $\delta
= Regress^*(c, \delta_G)$. We print a solution if we find a
plan-state pair $\langle c, \delta \rangle \in N$ such that
$\sigma_I \in ext(\delta)$ since $c$ would be one solution
(Theorem \ref{lem3}). Otherwise, we regress from $N_s$ (the set of
all the p-states occurring in $N$). This process involves the
regression using non-sensing actions and sensing actions which are
applicable in $N_s$. The algorithm will stop with failure if (i)
we cannot regress from $N_s$; or (ii) no new p-state can be found.
Below, we list the main steps of the algorithm:

\begin{algorithm}
{\bf Solve}(P)  where $ P  = \langle A,O,I,G \rangle$
\begin{list}{}{\topsep=2pt \itemsep=2pt \parsep=2pt}
\item [{\bf 1.}] Let $N = \{\langle [], \delta_G \rangle\}$ ($N_s
= \{\delta_G\}$).

\item [{\bf 2.}] {\bf Repeat}

\item [{\bf 3.}]   If there exists some $\langle c,\delta \rangle
\in N$ such that. $\sigma_I \in ext(\delta)$ then prints $c$ as a
solution.

\item [{\bf 4.}] Do one of the following:
\begin{list}{}{\topsep=2pt \itemsep=2pt \parsep=2pt}

\item [{\bf 4.1}] Find a $\langle c, \delta \rangle \in N$, a
non-sensing action $a$ such that. $a$ is applicable in $\delta$
and $\delta' = Regress(a,\delta) \not\in N_s$. Add $\langle a;c,
\delta' \rangle$ to $N$.

\item [{\bf 4.2}] Find a set $\Delta =
    \{\langle c_1, \delta_1 \rangle, \ldots,
    \langle c_n, \delta_n \rangle \} \subseteq N$, a sensing
    action $a$, and a set of formulas $\chi = \{\varphi_1,\ldots,\varphi_n\}$
    such that (i) $\chi$ spans over some $\emptyset \neq S_a \subseteq
    Sens_a$, (ii) $a$ is applicable in
    $\Gamma = \{\delta'_i \mid \delta'_i = [\delta_i.T \cup \varphi_i^+,
                            \delta_i.F \cup \varphi_i^-] \}$,
    and that (iii) $\delta' {=}
    Regress(a,\Gamma) {\not}{\in} N_s$. Add
    $\langle a;
    case(\varphi_1 {\rightarrow} c_1, \ldots, \varphi_n {\rightarrow}
    c_n), \delta' \rangle$
     to $N$.


\end{list}

\item [{\bf 5.}] {\bf Until} $N$ does not change.

\item [{\bf 6.}] Return NO SOLUTION.
\end{list}
\end{algorithm}

\COMMENT

\st The algorithm for step 4.2 is described in details as follows:
Given a set of p-states $N$ and a sensing action $a$ with the
sense list $S_a$, the algorithm for finding subsets of $N$ that
$a$ is applicable can be divided into two steps:

\begin{list}{$\bullet$}{\topsep=2pt \itemsep=2pt \parsep=2pt}
    \item partitioning $N$ into
        disjoint subsets $S_1, \ldots, S_n,$, such that
        $N = S_1 \cup \ldots \cup S_n$ and that for any two
        p-sates $s \in S_i, s' \in S_j (1 \leq i \neq j \leq n)$,
        there is no set of p-states $V$ that $s \in V, s' \in V$
        and $a$ is applicable in $V$;
    \item combining p-states of $S_i,\ (i=1,\ldots n)$, into sets regressable by $a$.
\end{list}

\st We briefly describe these two steps as follows: We say that
two p-states $s_1$ and $s_2$ agree on a fluent $f$ if $f$ is true
(resp. false) in $s_1$ implies that $f$ is true (resp. false) in
$s_2$ and vice versa. $s_1$ and $s_2$ agree on a set of fluents
$\Delta$ if they agree on every fluent belonging to $\Delta$.
Let's denote the set of fluents that do not belong to $S_a$ by
$\overline{S_a}$. For each $s \in N$, let $K(s)$ be the set of
p-states that agree with $s$ on the set $\overline{S_a}$. It is
easy to see that (i) if $s$ and $s'$ do not agree on the set
$\overline{S_a}$ then $K(s) \cap K(s') = \emptyset$; and that (ii)
if $s$ and $s'$ do not agree on the set $\overline{S_a}$ then they
cannot belong to a set of p-states that is regressable by $a$.
Therefore, we partition $N$ into $S_1, \ldots, S_n$ where each set
$S_i$ agrees on $\overline{S_a}$.

\st An example for the computation is given below:

\st Let $N = \{[\emptyset,pq], [\{p\},\{q\}], [\{q\},\{p\}],
[\{p,q\},\emptyset]\}$ and let $S_a = \{p\}$ then the computation
will partition $N$ into:

    \begin{list}{$\bullet$}{\topsep=2pt \itemsep=2pt \parsep=2pt}
        \item $K([\emptyset,\{p,q\}]) = \{[\emptyset,\{p,q\}], [\{p\},\{q\}]\}$)
        \item  $K([\{q\},\{p\}]) = \{[\{q\},\{p\}], [\{p,q\},\emptyset]\}$
    \end{list}

\st Clearly that if the result of the above computation is a
    sequence of sets of p-states $S_1, \ldots, S_n$ then for every
    $i=1,\ldots,n$, if $s$ and $s'$ belong to $S_i$, then $s$ and $s'$ agree on
    the set $\overline{S_a}$.

\st Let $S$ be a set of p-states that for each pair of $s$ and
$s'$ in $S$, $s$ and $s'$ agree on the set $\overline{S_a}$. We
now split $S$ into different sets of p-states such that $a$ is
applicable in these sets: We say that two p-states $s_1, s_2$
reflex over a fluent $f$ if $f$ is false (resp. true) in $s_1$,
and true (resp. false) in $s_2$, and that by making $f$ false
(resp. true) in $s_2$, we have $s_1$. For example $s_1 =
[\emptyset,pq]$ and $s_2=[p,q]$ reflex over $p$. We say two sets
of p-states $\{s_1, \ldots, s_n\}$ and $\{s'_1, \ldots, s'_n\}$
reflex over a fluent $f$ if $f$ is false (resp. true) in every
p-state $s_1, \ldots,s_n$, and true (resp. false) in every p-state
$s'_1, \ldots,s'_n$, and that by making $f$ false (resp. true) in
every p-state $s'_1, \ldots,s'_n$, we have $s_1, \ldots,s_n$.
Given any two sets of p-states $S_1 = \{s_1, \ldots,s_n\}$ and
$S_2 = \{s'_1, \ldots,s'_n\}$ such that $S_1$ and $S_2$ reflex
over a fluent $f$. If $a$ is regressable in $S_1$ with respect to
a set of fluents $S_a=\{f_1, \ldots,f_m\}$ and also regressable in
$S_2$ with respect to $S_a$ then $a$ is regressable in $S_1 \cup
S_2$ with respect to the set $\{f,f_1,\ldots,f_n\}$. This
discussion leads to the following steps for combining $S$ to sets
of regressable p-states:

        \begin{list}{$\bullet$}{\topsep=2pt \itemsep=2pt \parsep=2pt}
        \item for each pair $\{s,s'\} \subseteq S$ such that $s,s'$ reflex
        over a fluent $f \in S_a$ we create the set $V=\{s,s'\}$. We have that
        $a$ is applicable in $V$;

        \item for each pair $V,V'$ which are sets of p-states obtained from
        the previous step where $V$ and $V'$ reflex over $f \in S_a$.
        Let $V_a \subseteq S_a$ be the set of sense fluents that $a$ is regressable
        in $V$ and $V'$ with respect to $V_a$ and that $f \not \in V_a$. We create the set
        $V \cup V'$ of p-states for which $a$ is applicable and the
        set of regressable fluents contains two elements;

        \item the above step is repeated until the number of regressable
        fluents equals $|S_a|$ or no more pair can be created.
        \end{list}

\st Consider again the set $N = \{[\emptyset,pq], [\{p\},\{q\}],
[\{q\},\{p\}], [\{p,q\},\emptyset]\}$ and let $S_a = \{p,q\}$ be a
sense list of an action $a$. $N$ is also the only partition after
the partitioning phase. The result of computing regressable sets
for $a$ is illustrated as follows:

        \begin{list}{$\bullet$}{\topsep=2pt \itemsep=2pt \parsep=2pt}
        \item first step: we create sets $S_1 = \{[\emptyset,pq],[p,q]\}$,
            $S_2 = \{[q,p], [pq,\emptyset]\}$, $S_3 = \{[\emptyset,pq],
            [q,p]\}$, and $S_4 = \{[p,q], [pq,\emptyset]\}$;
        \item second step: Combine $S_1,S_2$ or $S_3,S_4$ and we have $N$ is
            regressable with respect to $\{p,q\}$.
        \end{list}

\ENDCOMMENT

\st The next theorem establishes the correctness of our
algorithm.%
\begin{theorem} \label{theo41}
For every $\langle c, \delta \rangle \in N$ where $N$ denotes the
set of plan-state pairs maintained by {\bf Solve}$(P)$,
$Regress^*(c,\delta_G) = \delta$.
\end{theorem}
\st {\textbf{Proof:} In Appendix.}

\st Since the algorithm searches through all possible regression
path, we have the following theorem.
\begin{theorem} \label{theo42}
For every planning problem $P = \langle A,O,I,G \rangle$,
\begin{enumerate}
\item {\bf Solve}$(P)$ will always stop;
\item if $P$ has a regression solution $c$ then {\bf Solve}$(P)$ will return a
conditional plan $c'$ such that $Regress(c,\delta_G) =
Regress(c',\delta_G)$; and
\item if $P$ has no regression solution
then {\bf Solve}$(P)$ will return NO SOLUTION.
\end{enumerate}
\end{theorem}
\st {\textbf{Proof:} In Appendix}

\st In the next example, we demonstrate how our algorithm works.

\begin{example} [Getting to Evanston -  con't]
Let us apply the algorithm to the problem of getting to Evanston.
Consider an initial condition $I= \{$ at{-}start, $\neg$
on{-}western, $\neg$ on{-}belmont, $\neg$ on{-}ashland, $\neg$
at{-}evanston, $\}$, and a goal condition $G =\{$ at-evanston$\}$.
So, $\delta_G = [\{$ at{-}evanston $\}$, $\{\}]$.
The algorithm goes through the following iterations: \\

\begin{figure}[h]
{\scriptsize
\begin{center}
\begin{tabular} {|l|l|l|}\hline\hline
\#I & Action ($a$) & Regressed-from member of $N$ \\ \hline
0 &  & $\langle [], \delta_G \rangle$\\
1 & $a_1 = $take{-}ashland  & $\langle [], \delta_G \rangle$ \\
2 & $b_1 = $take{-}western  & $\langle [], \delta_G \rangle$ \\
3 & $a_2 = $take{-}belmont & $\langle a_1, \delta1_1 \rangle$ \\
4 & $b_2 = $goto{-}western{-}at{-}belmont &
             $\langle b_1, \delta2_1 \rangle$ \\
5 & $a_3 = $goto{-}western{-}at{-}belmont &
             $\langle a_2;a_1, \delta1_2 \rangle$ \\
6 & check{-}traffic &  $\langle a_3;a_2;a_1, \delta1_3 \rangle$,
       $\langle b_2;b_1, \delta2_2 \rangle$ \\
\hline\hline \#I &
$Regress(a,\delta)$/$Regress(a,\{\delta_1,\ldots,\delta_n\})$
 & New member of $N$ \\ \hline
0 & &  \\
1 &       $\delta1_1 = [\{$on{-}ashland$\}, \{$$\}]$ &
          $\langle a_1, \delta1_1 \rangle$ \\
2 &       $\delta2_1 = [\{$on{-}western$\}, \{$traffic{-}bad$\}]$&
          $\langle b_1, \delta2_1 \rangle$ \\
3 &      $\delta1_2=[\{$on{-}belmont, traffic{-}bad$\}, \{$$\}]$ &
          $\langle a_2;a_1, \delta1_2 \rangle$ \\
4 &       $\delta2_2=[\{$at{-}start$\}, \{$traffic{-}bad$\}]$ &
          $\langle b_2;b_1, \delta2_2 \rangle$ \\
5 &       $\delta1_3=[\{$at{-}start,traffic{-}bad$\}, \{$$\}]$ &
          $\langle a_3;a_2;a_1, \delta1_3 \rangle$ \\
6 &       $\delta1_4 = [\{$at{-}start$\}, \{$$\}]$ &
          $\langle p,\delta1_4 \rangle$ \\
\hline
\end{tabular}
\end{center}
} \centerline{where $p= check$-$traffic; case(traffic$-$bad\rightarrow
a_3;a_2;a_1$, $\neg $ $traffic$-$bad\rightarrow b_2;b_1$).}
\caption{Algorithm illustration.}
    \label{example-fig} \hfill $\Box$
\end{figure}
\end{example}

\COMMENT
{\scriptsize
\begin{tabular}{|l|l|l|l|l|} \hline\hline
  & Action ($a$) & Regressed-from member of $N$ & $Regress(a,\delta)$/
$Regress(a,\{\delta_1,\ldots,\delta_n\})$
 & New member of $N$ \\ \hline
0 &              & $\langle [], \delta_G \rangle$  & & \\
1 & $a_1 = $take{-}ashland  & $\langle [], \delta_G \rangle$ &
          $\delta1_1 = [\{$on{-}ashland$\}, \{$$\}]$ &
          $\langle a_1, \delta1_1 \rangle$ \\
2 & $b_1 = $take{-}western  & $\langle [], \delta_G \rangle$ &
          $\delta2_1 = [\{$on{-}western$\}, \{$traffic{-}bad$\}]$ &
          $\langle b_1, \delta2_1 \rangle$ \\
3 & $a_2 = $take{-}belmont & $\langle a_1, \delta1_1 \rangle$ &
      $\delta1_2=[\{$on{-}belmont, traffic{-}bad$\}, \{$$\}]$ &
          $\langle a_2;a_1, \delta1_2 \rangle$ \\
4 & $b_2 = $goto{-}western{-}at{-}belmont &
             $\langle b_1, \delta2_1 \rangle$ &
      $\delta2_2=[\{$at{-}start$\}, \{$traffic{-}bad$\}]$ &
          $\langle b_2;b_1, \delta2_2 \rangle$ \\
5 & $a_3 = $goto{-}western{-}at{-}belmont &
             $\langle a_2;a_1, \delta1_2 \rangle$ &
      $\delta1_3=[\{$at{-}start,traffic{-}bad$\}, \{$$\}]$ &
          $\langle a_3;a_2;a_1, \delta1_3 \rangle$ \\
6 & check{-}traffic &  $\langle a_3;a_2;a_1, \delta1_3 \rangle$,
       $\langle b_2;b_1, \delta2_2 \rangle$ &
       $\delta1_4 = [\{$at{-}start$\}, \{$$\}]$ &
       $\langle p, \delta1_4 \rangle$
      \\
\hline
\end{tabular}
} where $p= \langle$check{-}traffic;
case(traffic{-}bad{$\rightarrow$} $a_3;a_2;a_1$, $\neg $
traffic{-}bad{$\rightarrow$} $b_2;b_1$), $\delta1_4 \rangle$.
\end{example}
\centerline{Figure 2: Algorithm illustration}

\subsection{Heuristics Control}

\st As \footnote{Implementation with heuristics is to be done.} we
employ the state space search approach, one of the common
techniques is using heuristics to efficiently guide the search. In
classical planning, the heuristics function over a state $S$ is
the cost estimate of a plan that achieves $s$ from the initial
state $S_0$. A state with optimal heuristics value is often chosen
to explore next. However, in the presence of sensing it is
necessary to select sets of states to regress over a sensing
action. The classical heuristics function thus needs to be
extended to cover this situation. In the following, we first
briefly discuss the heuristics function of $HSP$-$r$ planner~\cite{BG01}.
We then extend that function to define a heuristics
function for both non-sensing and sensing action.

\st $HSP$-$r$ follows the {\it Greedy Best Search} which uses the
cost function $f(S) = g(S) + w*h(S)$, where $g(S)$ is the number
of actions regressed from the goal state, $w$ is a weigh constant,
and $h(S)$ is the heuristics value of $S$. $h(S)$ is computed by
either taking the sum or max of the cost for each individual
fluent making up $S$. For each fluent $p \in S$ and for each
action $a$ that adds $a$ to $S$, $h(p)$ is iteratively updated to
a fixpoint as $h(p) \leftarrow min\{h(p),1+h(Prec(a))\}$ where
$Prec(a)$ is the precondition of $a$. Initially, each proposition
$p$ is assigned the cost $\infty$ if $p$ does not appear in the
initial state; otherwise it is assigned value $0$.

\st Given a p-state $\delta$ and a set of p-states $\Delta$ our
extended heuristics function is defined as follows:

\begin{itemize}
    \item non-sensing action:

    We assume that achieving a positive fluent and achieving a
    negative fluent are independent from each other. This leads to
    the following heuristics:
    \begin{definition} [Sum Heuristics] \label{sum} $h_{sum}(\delta) \leftarrow
    \sum_{p \in \delta.T} h(p) + \sum_{q \in \delta.F} h(q)$
    \end{definition}

    The value $h(p)$ (resp. $h(q)$) is computed as follows:
    Initially, each proposition $p$ is assigned the cost $\infty$ if
    $p$ does not appear in the initial state $\sigma_I$; otherwise it is assigned
    value $0$. For a non-sensing action $a$ applicable in $\delta$
    and that $p \in Add_a$ ($q \in Del_a$):
    $h(p) \leftarrow min\{h(p), 1 + h(Pre_a)\}$ (resp.
    $h(q) \leftarrow min\{h(q), 1 + h(Pre_a)\}$), where $h(Pre_a)$
    is computed using Definition (\ref{sum}).

    To account for the fact that achieving a positive fluent and achieving a
    negative fluent are not independent from each other, we have
    the following heuristics:
    \begin{definition} [Max Heuristics] \label{max} $h_{max}(\delta) \leftarrow
    max_{p \in \delta.T \cup \delta.F} h(p)$
    \end{definition}

    \item sensing action:

    For a sensing action $a$ applicable in $\Delta$, let $\delta' =
    Regress(a,\Delta)$. As we can always reach the set $\Delta$ from the
    $\delta'$ by executing $a$ in $\delta'$, the
    heuristics for $\Delta$ can be computed as follows:
    \begin{definition} [Sensing Heuristics] \label{sense} $h_{sense}(\Delta) \leftarrow
    1 + h(\delta')$
    \end{definition}

\end{itemize}

\st Note that, if there exists no action $a$ applicable in
$\Delta$ then $h(\Delta) = \infty$. As the heuristics function
$h(\delta)$ and $h(\Delta)$ have been defined, we also need to
define the function $g(\delta)$ and $g(\Delta)$.

\begin{definition} [Accumulated Cost] \label{acmCost} For a p-state
$\delta$ and a set of p-states $\Delta$:
\begin{itemize}
    \item $g(\delta) = 0$ if $\delta$ is the goal state $\delta_G$;
    \item $g(\delta) = min\{C1, \ldots, C_n\}$ where $C_i$ is
        the total number of actions of a plan $i$ leading
        from $\delta$ to the goal state; \footnote{Computing
        $g(\delta)$ in this case is straightforward as each state
        in our planning algorithm contains a plan leading from
        that state to the goal.}
    \item $g(\Delta) = \sum_{\delta' \in \Delta} g(\delta')$.
\end{itemize}
\end{definition}

\begin{definition} [Cost Function] \label{costF} For a p-state
$\delta$ and a set of p-states $\Delta$, a cost function is
defined as follows: $$f(\delta) = g(\delta) + w* h(\delta)$$
$$f(\Delta) = g(\Delta) + w* h(\Delta)$$ where $w$ is a positive
constant.
\end{definition}

\st It's easy to see that a p-state has a finite heuristics value
if there exists a finite number of non-sensing actions that lead
from that p-state to one of goal states in $\Delta_I$. A set of
p-states has a finite heuristics value if their regressed p-state
has a finite heuristics value. We may get the first plans faster
by modifying the step 4 in the planning algorithm to produce only
p-states that have a finite heuristics value as soon as that
p-states appear.

\st When regression involves sensing actions, a p-state obtained
after the algorithm loops $k$ times may be combined with a p-state
obtained after the algorithm loops only one time to form a new
regressed p-state having a finite heuristics value. If $k$ is
large then finding all plans of length $k+1$ is costly.

\ENDCOMMENT

\st We now describe our initial experiments in the next section.

\section{Experimentation} \label{expr}

We have experimentally compared our system with the two systems
\cite{SGP98,STB03} in domains with {\em sensing actions and
incomplete information} but did not compare our planner with
\cite{PC96} since the planner in \cite{SGP98} is significantly
better than that of \cite{PC96}. We also did not compare our
system with others that deal with nondeterministic or
probabilistic actions as our action representation does not have
this capability.

\st We run our Java-based planner with three well known {\it
domains with sensing actions:} {\it Cassandra, Bomb in the toilet,
and Sickness domain.} These domains are obtained from the SGP
distribution \cite{SGP98}. All experiments are run on a Compaq
laptop 1.8Ghz CPU with 512 MbRAM. The experimental result
(obtained without using heuristics) is presented in Figure (\ref{example-fig05}).
It is necessary to note that, Figure (\ref{example-fig05}) is a crude comparison as
the other two use static causal laws and boolean sensing fluents
(e.g. in Bomb in the toilet domain) while ours uses multi-valued
sensing fluents; and the Logic Programming based planner
($\pi(P)$) uses conditional effects but ours does not.

\begin{figure}[h]
    \centerline{
\scriptsize
\begin{tabular}{|l|c|c|c|c|c|} \hline \hline
\multicolumn{1}{|l|} {Domains/} & \multicolumn{5}{|c|} {Planners
(time in milliseconds)} \\ \cline{2-6}
\multicolumn{1}{|l|} {Problem} & \multicolumn{3}{|c|} {aSense} &
\multicolumn{1}{|c|} {$\pi(P)$} & \multicolumn{1}{|c|} {SGP}\\
\cline{2-4}
\multicolumn{1}{|l|} {} & \multicolumn{1}{|c|} {preprocessing} &
\multicolumn{1}{|c|} {search} & \multicolumn{1}{|c|} {total} &
\multicolumn{1}{|c|} {} & \multicolumn{1}{|c|} {}\\ \hline \hline
{\bf Cassandra} &  & & & & \\
a1-prob & 50 & 10 & {\bf 60} & 510 & 130\\
a2-prob & 50 & 10 & {\bf 60} & 891 & 60\\
a3-prob & 70 & 0 & {\bf 70} & 119 & 70\\
a4-prob & 60 & 220 & {\bf 280} & 1030 & 431\\
a5-prob & 30 & 10 & {\bf 40} & 130 & 20\\
a6-prob & 200 & 1392 & {\bf 1592} & 18036 & NA \footnote{Memory problem}\\
a7-prob & 40 & 10 & {\bf 50} & 150 & 110\\ \hline \hline
{\bf Bomb} & && &&\\
bt-1sa & 40 & 0 & {\bf 40} & 15812 & 751\\
bt-2sa & 40 & 10 & {\bf 50} & 18676 & 1161\\
bt-3sa & 40 & 10 & {\bf 50} & 18445 & 1512\\
bt-4sa & 200 & 10 & {\bf 210} & 22391 & 1892\\ \hline
\end{tabular}
}
\normalsize
\caption{Running time for the Cassandra and Bomb In The Toilet domains.}
\label{example-fig05}
\end{figure}


\COMMENT

\section{Related Work} \label{relate}

Regression with respect to simple actions has been studied in e.g.
\cite{Ped94,Rei98,BG01}. Regression with respect to sensing
actions has been studied in e.g. \cite{SL93,SL03,SB01}. In the
planning literature there has been a lot of work
\cite{PS92,Eetal92,GB94,PC96,SGP98,L98,CRT98,BG00,R00,R02,STB03}
in developing planners that generate conditional plans in presence
of incomplete information, some of which use sensing actions and
the others do not.

Our work in this paper is related to different approaches to
regression and planning in the presence of sensing actions and
incomplete information. It differs from earlier formula regression
such as \cite{Ped94,Rei98,SB01} in that it is a state-based
formulation and the other are formula based. Unlike the
conditional planners \cite{PS92,CRT98}, our planner can deal with
sensing actions similar to the planners in
\cite{Eetal92,L98,STB03,SGP98}. However, it does not deal with
nondeterministic and probabilistic actions such as the planners in
\cite{BG00,PC96,R00,R02}. It is also not a conformant planner as
in \cite{CRT98,eit00}. For these reasons, we currently compare our
planner with those of \cite{STB03,SGP98}.

\ENDCOMMENT

\section{Conclusion and Future Work} \label{conc}

\st In this paper, we used the 0-approximation semantics
\cite{SB01} and defined regression with respect to that semantics.
We considered domains where an agent does not have complete
information about the world, and may have sensing actions. We
first started with domains having only Boolean fluents and
formally related our definition of regression with the earlier
definition of progression in \cite{SB01}. We showed that planning
using our regression function would not only give us correct plans
but also would not miss plans. We then presented a search
algorithm for generating conditional plans. Lastly, we presented
preliminary experimental results and discussed difficulties we
faced as well as future enhancements. To simplify our formulation,
we used the STRIPS-like action representation and considered
fluents with finite domains.

\st Our planner is sound, however the use of the computationally
less complex 0-approximation leads to incompleteness with respect
to the full semantics. 
This is a trade-off to counter the higher complexity thus leading
to the efficiency in search for plans. Other limitations due to
state space regression are difficulties in handling static causal
laws and conditional effects. To further improve the search
efficiency, we plan to develop necessary heuristics by extending
the work of \cite{BG01} to handle sensing actions. We also plan to
extend our results to non-binary domains. Lastly, we need to
directly consider actions with conditional effects,
nondeterministic actions, and static causal laws and develop
regression operators for these cases.

\COMMENT

\st In this paper, we used the 0-approximation
semantics~\cite{SB01} and defined regression with respect to that
semantics. We considered domains where an agent does not have
complete information about the world, and may have sensing
actions. We first started with domains having only Boolean fluents
and formally related our definition of regression with the earlier
definition of progression in~\cite{SB01}. We showed that planning
using our regression function would indeed give us correct plans.
We then presented a search algorithm for generating conditional
plans. Next, we considered domains with multi-valued fluents and
extended our approach to these domains. Lastly, we presented
preliminary experimental results and discussed difficulties we
faced as well as future enhancements. To simplify our formulation,
we used the STRIPS-like action representation and considered
fluents with finite domains.

\st Our planner is sound, however the use of 0-approximation leads
to its incompleteness. This is a trade-off to counter the higher
complexity thus leading to the efficiency of search for plans.
Other limitations due to state space regression are difficulties
in handling static causal laws and conditional effects. An issue
that needs to be addressed in the future is to find sufficiency
conditions that would guarantee the completeness of our approach
with respect to the 0-approximation. We also need to directly
consider actions with conditional effects, nondeterministic
actions,  and static causal laws and develop regression operators
for these cases.

\ENDCOMMENT


\section*{APPENDIX}

\setcounter{section}{3}

\setcounter{theorem}{3}

{\lemma [Sensed Set] \label{lemSS} Consider a sensing action $a$
and a set of p-states $\Delta$. If there exists a sensed set of
$\Delta$ with respect to $a$ then it is unique.}

\begin{proof} [Lemma \ref{lemSS}] Assume that $X$ and $X'$ are two
different sensed sets of $\Delta$ with respect to $a$. Since $X
\neq \emptyset$, let's consider a fluent $f \in X$. By Definition
\ref{sensed-Set}, for two partitions $(\{f\}, X \setminus \{f\})$
and $(X \setminus \{f\}, \{f\})$ of $X$, there exist $\delta_i \in
\Delta$ and $\delta_j \in \Delta$ ($1 \leq i \neq j \leq n$) such
that $\{f\} = \delta_i.T \cap X$ and $\{f\} = \delta_j.F \cap X$,
i.e. $f$ is true in $\delta_i$ and false in $\delta_j$ [*].

\st Suppose that $f \not \in X'$. By Definition \ref{sensed-Set},
we must have that: either $f \in \delta_k.T \setminus X'$ or $f \in
\delta_k.F \setminus X'$ for all $k$, $1 \leq k \leq n$, i.e $f$ is either
true or false in every $\delta_k \in \Delta$. In either case, this
contradicts with [*]. Therefore, $f \in X'$.

\st Similarly, we can argue that, if $f \in X'$ then $f \in X$.
Thus, $f \in X$ iff $f \in X'$, i.e. $X=X'$.
\end{proof}

\setcounter{section}{3}

\setcounter{theorem}{6}

{\lemma [Unique Sensed Set] \label{lemSS1} Consider a sensing
action $a$ and a set of p-states $\Delta$ such that $a$ is
applicable in $\Delta$. Let $\Delta' {=} \{\delta'_1, \ldots,
\delta'_n\}$, where $\delta'_i$ is a partial extension of
$\delta_i$ ($i=1, \ldots,n$), $\Delta'' {=} \{{\delta_1}'',
\ldots, {\delta_n}''\}$, where ${\delta_i}''$ is a partial
extension of $\delta_i$ ($i=1, \ldots,n$). If $p(a,\Delta') \neq
\bot$ and $p(a,\Delta'') \neq \bot$ then $p(a,\Delta') =
p(a,\Delta'')$.}

\begin{proof} [Lemma \ref{lemSS1}] Assume that $p(a,\Delta') \neq
p(a,\Delta'')$. Since $p(a,\Delta') \neq \bot$, there exists $f
\in p(a,\Delta')$ where $f \not \in p(a,\Delta'')$.


\st Since $\Delta''$ is proper with respect to $p(a,\Delta'')$,
$f \in Sens_a$, and $f$ is known in $\Delta''$, by Definition
\ref{sensed-Set}, we must have that either (i) $f \in
{\delta_i}''.T \setminus p(a,\Delta'')$ for every $1 \leq i \leq
n$, or (ii) $f \in {\delta_i}''.F \setminus p(a,\Delta'')$ for
every $1 \leq i \leq n$.

\st Consider case (i). We have that $f \in {\delta_i}''.T$ for all
$1 \leq i \leq n$ [*].

\st Since $f \in p(a,\Delta')$, by Definition \ref{sensed-Set},
for the partition $(p(a,\Delta') \setminus \{f\}, \{f\})$ of
$p(a,\Delta')$, there exists $\delta'_j \in \Delta'$ ($1 \leq j
\leq n$) such that $\delta'_j.F \cap p(a,\Delta') = \{f\}$, i.e.
$f$ is false in $\delta'_j$. Since $\delta'_j$ is a partial
extension of $\delta_j$, we have that $\delta_j.F \subseteq
\delta'_j.F$. Also, as $Sens_a$ is known in $\delta_j$, we must
have that $f \in \delta_j.F$. Since ${\delta_j}''$ is also a
partial extension of $\delta_j$, we have that $\delta_j.F
\subseteq {\delta_j}''.F$, therefore $f \in {\delta_j}''.F$. From
[*], we also have $f \in {\delta_j}''.T$. This is a contradiction.

\st Similarly, we can show a contradiction for case (ii).
Therefore, we conclude that if $f \in p(a,\Delta')$ then $f \in
p(a,\Delta'')$. Using similar arguments, we can also show that for
any $f \in p(a,\Delta'')$, $f \in p(a,\Delta')$. Therefore,
$p(a,\Delta') = p(a,\Delta'')$. \end{proof}

\setcounter{section}{8}

\setcounter{theorem}{0}

{\lemma \label{lem0-appdx} Let $\sigma'$ be an a-state and $a$ be
a sensing action executable in $\sigma'$. For any $S_a \subseteq
Sens_a$ and $\sigma \in \Phi(a,\sigma')$, let $\sigma.T \cap S_a =
S^+_\sigma$ and $\sigma.F \cap S_a = S^-_\sigma$, we have that
$S^+_\sigma \cup S^-_\sigma = S_a$ and $S^+_\sigma \cap S^-_\sigma
= \emptyset$.}

\begin{proof} [Lemma \ref{lem0-appdx}] It is easy to see that the
lemma is correct for the case $S_a = \emptyset$. Let's consider
the case $S_a \neq \emptyset$. Since $S^+_\sigma \subseteq
\sigma.T$ and $S^-_\sigma \subseteq \sigma.F$, we have that
$S^+_\sigma \cap S^-_\sigma = \emptyset$.

Consider $f \in S^+_\sigma \cup S^-_\sigma$, we have that $f \in
S^+_\sigma$ or $f \in S^-_\sigma$. In both cases, we have $f \in
S_a$.

Consider $f \in S_a$. Since $S_a \subseteq Sens_a$, we have that
$f \in Sens_a$. By the definition of $\Phi$, we have that $f \in
\sigma.T$ or $f \in \sigma.F$. From this fact, it's easy to see
that $f \in S^+_\sigma$ or $f \in S^-_\sigma$. \end{proof}

{\lemma \label{lem1-appdx} Let $\delta$ be a p-state. An a-state
$\sigma$ is an extension of $\delta$ (i.e. $\sigma \in
ext(\delta)$) iff $\sigma$ is an a-state of the form $\langle
\delta.T \cup X, \delta.F \cup Y \rangle$ where $X, Y$ are two
disjoint sets of fluents and $X \cap \delta.F = \emptyset,\ Y \cap
\delta.T = \emptyset$.}

\begin{proof} [Lemma \ref{lem1-appdx}]:

\begin{itemize}
    \item [$\bullet$] Case ``$\Rightarrow$'':

    Let $\sigma \in ext(\delta)$ be an extension of $\delta$. By
    the definition of an extension, $\sigma$ is an a-state where $\delta.T \subseteq
    \sigma.T$ and $\delta.F \subseteq \sigma.F$. Denote $X=\sigma.T \setminus
    \delta.T$ and $Y = \sigma.F \setminus \delta.F$. Clearly, $X$
    and $Y$ are two set of fluents where $X \cap Y = \emptyset$ and
    $X \cap \delta.F = \emptyset,\ Y \cap \delta.T = \emptyset$.
    \item [$\bullet$] Case ``$\Leftarrow$'':

    Let $\sigma$ be an a-state of the form $\langle \delta.T \cup
    X, \delta.F \cup Y \rangle$ where $X, Y$ are two disjoint
    sets of fluents and $X \cap \delta.F = \emptyset,\ Y \cap \delta.T
    = \emptyset$.

    It's easy to see that $\sigma.T \cap \sigma.F = \emptyset$,
    i.e. $\sigma$ is consistent. Furthermore, $\delta.T \subseteq
    \sigma.T$ and $\delta.F \subseteq \sigma.F$, i.e. by definition
    of an extension, $\sigma$ is an extension of $\delta$.
\end{itemize}
\end{proof}

\setcounter{section}{3}

\setcounter{theorem}{8}

{\proposition [Non-sensing action] \label{lem1} Let $\delta$ and
$\delta'$ be two p-states, and $a$ be a non-sensing action. If
$Regress(a,\delta)= \delta'$ and $\delta' \neq \bot$, then for
every $\sigma'' \in ext(\delta')$ we have that (i) $\bot \not \in
\Phi(a, \sigma'')$, and (ii) $\Phi(a, \sigma'') \subseteq
ext(\delta)$.}

\begin{figure}[h]
    \centerline{
    \begin{tabular}{c}
    {
    \epsfxsize=3in \epsfbox{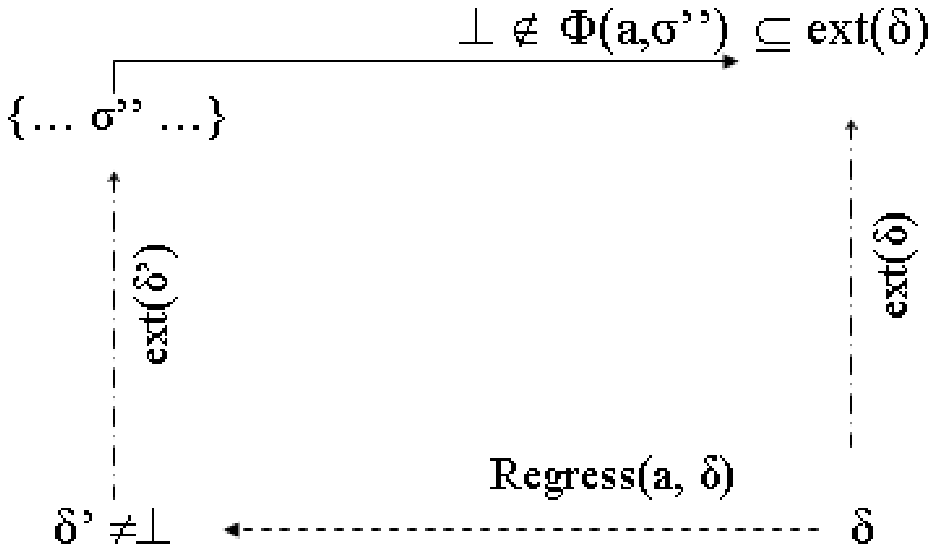}
    }
    \end{tabular}}
    \caption{Illustration of Proposition \ref{lem1}.}
    \label{example-fig1}
\end{figure}

\begin{proof} [Proposition \ref{lem1}] Let $\delta = [T,F]$. From
the fact that $Regress(a,\delta)= \delta' \neq \bot$, we have that
$a$ is applicable in $\delta$.

\st By Definition \ref{reg-nonsensing},
\[
\delta' = Regress(\delta,a) = [T \setminus Add_a \cup Pre^+_a, F
\setminus Del_a \cup Pre^-_a].
\]

\st Let $\sigma'' \in ext(\delta')$, we will show that (i) $\bot
\not \in \Phi(a, \sigma'')$ and (ii) $\Phi(\sigma'', a) \subseteq
ext(\delta)$.

\st Indeed, it follows from Lemma \ref{lem1-appdx} that
\[
\sigma'' = \langle (T \setminus Add_a) \cup Pre^+_a \cup X, (F
\setminus Del_a) \cup Pre^-_a \cup Y \rangle,
\]
where $X$ and $Y$ are two sets of fluents such that $\sigma''.T
\cap \sigma''.F = \emptyset$. We now prove (i) and (ii).

\begin{itemize}
    \item [$\bullet$] Proof of (i):

Since $Pre^+_a \subseteq \sigma''.T$ and $Pre^-_a \subseteq
\sigma''.F$, we conclude that $a$ is executable in $\sigma''$,
i.e. $\bot \not \in \Phi(a, \sigma'')$.

    \item [$\bullet$] Proof of (ii):

\st By definition of the transition function $\Phi$, we have that
\[
\Phi(a, \sigma'') = \{\langle ((T \setminus Add_a) \cup Pre^+_a
\cup X) \setminus Del_a \cup Add_a, ((F \setminus Del_a) \cup
Pre^-_a \cup Y) \setminus Add_a \cup Del_a \rangle\}
\]

\st Since $a$ is applicable in $\delta$, we have that $T \cap
Del_a = \emptyset$, $F \cap Add_a = \emptyset$. Furthermore,
$Del_a \cap Add_a = \emptyset$. Therefore, we have that $((T
\setminus Add_a) \cup Pre^+_a \cup X) \setminus Del_a \cup Add_a =
(T \setminus Add_a) \cup ((Pre^+_a \cup X) \setminus Del_a) \cup
Add_a \supseteq T \cup ((Pre^+_a \cup X) \setminus Del_a)
\supseteq T$. This concludes that $T \subseteq \Phi(a,
\sigma'').T$. Similarly, we have that $F \subseteq \Phi(a,
\sigma'').F$. This shows that $\Phi(a, \sigma'') \subseteq
ext(\delta)$.
\end{itemize}
\end{proof}

{\proposition [Sensing action] \label{lem2} Let
$\Delta=\{\delta_1, \ldots, \delta_n\}$ be a set of p-states,
$\delta'$ be a p-state, and $a$ be a sensing action. If
$Regress(a,\Delta) = \delta'$, where $\delta' \neq \bot$, then for
every $\sigma'' \in ext(\delta')$, we have that (i) $\bot \not \in
\Phi(a, \sigma'')$, and (ii) $\Phi(a,\sigma'') \subseteq
ext(\delta_1) \cup \ldots \cup ext(\delta_n)$.}

\begin{figure}[h]
    \centerline{
    \begin{tabular}{c}
    {
    \epsfxsize=3.5in \epsfbox{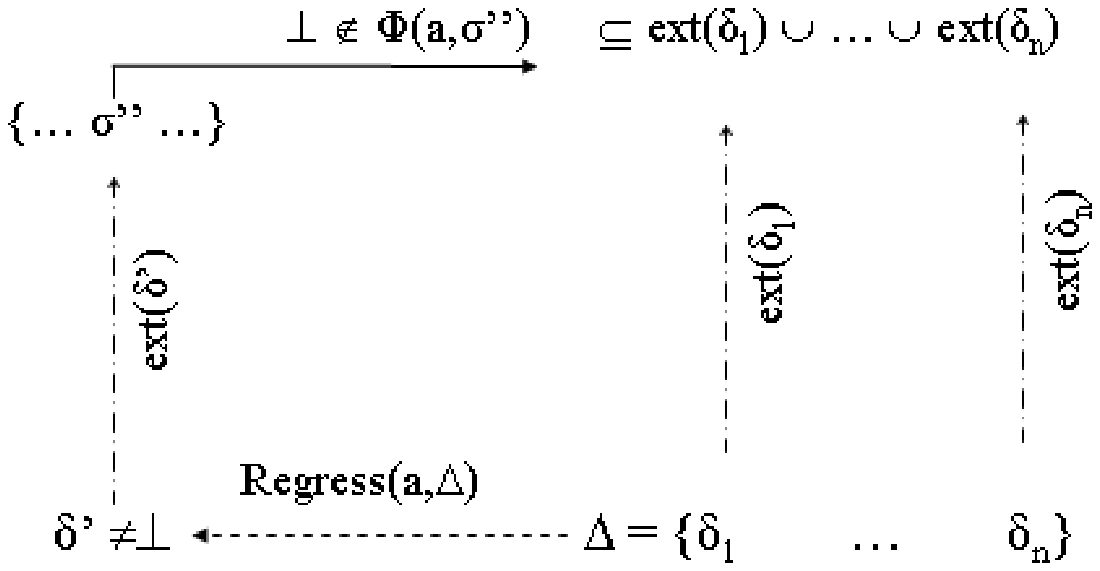}
    }
    \end{tabular}}
    \caption{Illustration of Proposition \ref{lem2}.}
    \label{example-fig1}
\end{figure}

\begin{proof} [Proposition \ref{lem2}] From the fact that
$Regress(a,\Delta) = \delta' \neq \bot$, we have that $a$ is
applicable in $\Delta$ with respect to some set $S_{a,\Delta}
\subseteq Sens_a$ ($S_{a,\Delta} \neq \emptyset$).

\st By Definition \ref{reg-sensing} we have:

\[ \delta' = Regress(a,\Delta) = [
(\bigcup^n_{i=1}\delta_i.T \setminus S_{a,\Delta}) \cup Pre^+_a,
(\bigcup^n_{i=1}\delta_i.F \setminus S_{a,\Delta}) \cup Pre^-_a ].
\]

\st Let $\sigma'' \in ext(\delta')$ be an arbitrary extension of
$\delta'$. We will show that (i) $\bot \not \in \Phi(a, \sigma'')$
and (ii) $\Phi(a,\sigma'') \subseteq ext(\delta_1) \cup \ldots
\cup ext(\delta_n)$.

\begin{enumerate}
    \item Proof of (i):

    It follows from Lemma \ref{lem1-appdx} that:

\[
\sigma'' = \langle (\bigcup^n_{i=1}\delta_i.T \setminus
S_{a,\Delta}) \cup Pre^+_a \cup X, (\bigcup^n_{i=1}\delta_i.F
\setminus S_{a,\Delta}) \cup Pre^-_a \cup Y \rangle
\]

\st where $X$ and $Y$ are two sets of fluents such that
$\sigma''.T \cap \sigma''.F = \emptyset$.

\st From the fact that $Pre^+_a \subseteq \sigma''.T$ and
 $Pre^-_a \subseteq \sigma''.F$, we conclude that $a$ is executable in
$\sigma''$, i.e. $\bot \not \in \Phi(a, \sigma'')$ [*].

\item Proof of (ii):


We need to prove that: for every $\sigma \in \Phi(a,\sigma'')$,
then there exists $\delta_i$ ($1 \leq i \leq n$) such that $\sigma
\in ext(\delta_i)$.

Indeed, consider an arbitrary $\sigma \in \Phi(a,\sigma'')$. Let's
denote $\sigma.T \cap S_{a,\Delta} = S^+_\sigma$ and $\sigma.F
\cap S_{a,\Delta} = S^-_\sigma$. By Lemma \ref{lem0-appdx}, we
have that $S^+_\sigma \cup S^-_\sigma = S_{a,\Delta}$ and
$S^+_\sigma \cap S^-_\sigma = \emptyset$.

Since $a$ is applicable in $\Delta$ with respect to
$S_{a,\Delta}$, by Definition \ref{app-sensing} and the definition
of $S_{a,\Delta}$, there exists $\Delta' = \{\delta'_1, \ldots,
\delta'_n\}$ where $\delta'_i$ is a partial extension of
$\delta_i$ ($1 \leq i \leq n$) such that $a$ is strongly
applicable in $\Delta'$ with respect to $S_{a,\Delta}$. By
Definition \ref{sensed-Set}, there exists $\delta'_i$ ($1 \leq i
\leq n$) such that $\delta'_i.T \cap S_{a,\Delta} = S^+_\sigma$
and $\delta'_i.T \cap S_{a,\Delta} = S^+_\sigma$. We will now show
that $\sigma \in ext(\delta_i)$ or in other word $\delta_i.T
\subseteq \sigma.T$ and $\delta_i.F \subseteq \sigma.F$.

Since $\delta_i.T \subseteq \delta'_i.T$, we have that $\delta_i.T
\cap S_{a,\Delta} \subseteq \delta'_i.T \cap S_{a,\Delta} =
S^+_\sigma$. Therefore:
\[\delta_i.T = \delta_i.T \setminus (\delta_i.T \cap S_{a,\Delta}) \cup
(\delta_i.T \cap S_{a,\Delta}) = (\delta_i.T \setminus
S_{a,\Delta}) \cup ( \delta_i.T \cap S_{a,\Delta}) \subseteq
(\delta_i.T \setminus S_{a,\Delta}) \cup S^+_\sigma.\]

Similarly, we can show that $\delta_i.F \subseteq (\delta_i.F
\setminus S_{a,\Delta}) \cup S^-_\sigma$.

Since $\sigma \in \Phi(a, \sigma'')$, by the definition of $\Phi$,
we have that $\sigma''.T \subseteq \sigma.T$. Let $\sigma.T
\setminus \sigma''.T = \omega$, we have that \[\sigma.T =
\sigma''.T \cup \omega = (\bigcup^n_{j=1}\delta_j.T \setminus
S_{a,\Delta}) \cup Pre^+_a \cup X \cup \omega.\] Since $\sigma.T
\cap S_{a,\Delta} = S^+_\sigma$ and $((\bigcup^n_{j=1}\delta_j.T
\setminus S_{a,\Delta}) \cup Pre^+_a) \cap S_{a,\Delta} =
\emptyset$ (because $Sens_a \cap Pre^+_a = \emptyset$), we must
have that $(X \cup \omega) \cap S_{a,\Delta} = S^+_\sigma$, i.e.
$S^+_\sigma \subseteq X \cup \omega$. From the fact that
$\delta_i.T \subseteq \delta_i.T \setminus S_{a,\Delta} \cup
S^+_\sigma$ and $S^+_\sigma \subseteq X \cup \omega$, it's easy to
see that $\delta_i.T \subseteq \sigma.T$. Similarly, we can show
that $\delta_i.F \subseteq \sigma.F$. From this fact, we conclude
that $\sigma \in ext(\delta_i)$ [**].

\end{enumerate}

From [*] and [**] the proposition is proved. \end{proof}

\setcounter{section}{8}

\setcounter{theorem}{0}

\begin{definition} [Branching Count] Let $c$ be a conditional plan,
we define the number of case plans of $c$, denoted by $count(c)$,
inductively as follows:
\begin{enumerate}
    \item if $c=[\ ]$ then $count(c) = 0$;
    \item if $c = a$, $a$ is a non-sensing action, then $count(c) = 0$;
    \item if $c_1$ and $c_2$ are conditional plans then $count(c_1;c_2) = count(c_1) + count(c_2);$
    \item if $c$ is a case plan of the form
        a; case($\varphi_1 \rightarrow c_1, \ldots, \varphi_n \rightarrow c_n $)
        where $a$ is a sensing action, then
        $count(c) = 1 + \sum^n_{i=1} count(c_i).$
\end{enumerate}
\end{definition}

{\observation \label{ob1} We have the following two observations:
\begin{enumerate}
    \item by Definition \ref{defC}, a conditional plan $c$ is a sequence of conditional plans $c_1;
    \ldots;c_n$ where (i) $c_i$ is either a sequence of non-sensing
    actions, or a sensing action followed by a case statement;
    and (ii) for every $i < n$, if $c_i$ is a sequence of
    non-sensing actions then $c_{i+1}$ is a case plan.
    \item let $\delta$ be a p-state, $\sigma$ be an
    extension of $\delta$, and $\varphi$ be a fluent
    formula. Then, $\delta \models \varphi$ implies $\sigma \models
    \varphi$.
\end{enumerate}
}

\setcounter{section}{8}

\setcounter{theorem}{2}

{\lemma \label{lem30} Let $\delta$ be a p-state and $c$ be a
conditional plan. Then, $Regression^*(c, \delta)$ is either a
p-state or $\bot$.}

\begin{proof} [Lemma \ref{lem30}]

The proof is done inductively over $count(c)$. The base case, c is
a sequence of non-sensing actions, follows immediately from items
1,2,4 of the $Regression^*$ definition (Definition \ref{d3}). The
inductive step follows from inductive hypothesis and the items 2,3
of the $Regression^*$ definition. \end{proof}

{\corollary [Sequence of Non-sensing action] \label{cor1} For
p-states $\delta$ and $\delta'$, and a sequence of non-sensing
actions $c = a_1; \ldots; a_n$ $(n \geq 1)$. $Regress^*(c,\delta)=
\delta' \neq \bot$ implies that $\bot \not \in \Phi^*(c,\sigma'')$
and $\Phi^*(c,\sigma'') \subseteq ext(\delta)$ for every $\sigma''
\in ext(\delta')$.}

\begin{proof} [Corollary \ref{cor1}] We prove the corollary by
induction over $\mid c \mid$, the number of non-sensing actions of
$c$.

\begin{itemize}
    \item [$\bullet$] Base case: $\mid c \mid = 1$

        This means that $c$ has only one action $a$. Using the
        Proposition \ref{lem1}, and Definition \ref{d3} -- item 2 -- the based case is
        proved. Notice that for the case $|c| = 0$, i.e. $c=[\ ]$,
        the corollary follows directly from Definitions \ref{d3}
        and \ref{def-extran}.
    \item [$\bullet$] Inductive Step:

        Assume that the corollary is shown for $\mid c \mid \leq k$ $(k \geq 1)$.
        We now prove the corollary for $\mid c \mid = k+1$.

        Let $c=a_1;\ldots; a_{k+1}$, and $c'=a_2;\ldots; a_{k+1}$
        where $a_i$ is a non-sensing action for $(1 \leq i \leq k+1)$.
        We have that $\mid c' \mid = k$. By Definition \ref{d3}

        \[
        Regress^*(c,\delta) = Regress(a_1, Regress^*(c',\delta)) = \delta'.
        \]

        \st Denote $Regress^*(c',\delta) = \delta^*$. Since $Regress(a_1, \delta^*) = \delta' \neq \bot$,
        we have that $\delta^* \neq \bot$.

        \st Let $\sigma'' \in ext(\delta')$. Since $\sigma'' \in
        ext(\delta')$, by Proposition \ref{lem1}, we have that $ \bot \not \in \Phi(a_1,\sigma'') = \{\sigma\} \subseteq
        ext(\delta^*)$, i.e. $\sigma \in ext(\delta^*)$.

        \st By the definition of $\Phi^*$, we also have that $\Phi^*(c, \sigma'') = \Phi^*(c', \Phi^*(a_1,\sigma''))$.
        Using the induction hypothesis for $|c'| = k$, where $Regress^*(c',\delta) = \delta^*$ and $\sigma \in ext(\delta^*)$, we have:

        \[
        \bot \not \in \Phi^*(c', \Phi^*(a_1, \sigma'')) = \Phi^*(c', \sigma) \subseteq ext(\delta).
        \]

        Therefore, $\bot \not \in \Phi^*(c,\sigma'')$ and $\Phi^*(c,\sigma'') \subseteq ext(\delta)$.

\end{itemize}
\end{proof}

{\lemma \label{lem30b} Let $\delta$ be a p-state and $c$ be a
conditional plan. If $Regress^*(c, \delta) = \delta'$ where
$\delta' \neq \bot$, then for every $\sigma \in ext(\delta')$: (i)
$\bot \not \in \Phi^*(c, \sigma)$ and (ii) $\Phi^*(c, \sigma)
\subseteq ext(\delta)$.}

\begin{proof} [Lemma \ref{lem30b}] We prove by induction on
$count(c)$, the number of case plans in $c$.
\begin{itemize}
    \item [$\bullet$] Base Case: $count(c)=0$. Then $c$ is a sequence of
    non-sensing actions. The base case follows from Corollary
    \ref{cor1}.
    \item [$\bullet$] Inductive Step: Assume that we have proved the
    lemma for $count(c) \leq k$ ($k \geq 0$). We need to prove the lemma for
    $count(c)=k+1$.

    Indeed, from Observation \ref{ob1}, let $c= c_1; \ldots; c_n$. By construction of $c$, we have two cases
    \begin{enumerate}
        \item $c_n$ is a case plan:

\begin{figure}[h]
    \centerline{
    \begin{tabular}{c}
    {
    \epsfxsize=3.7in \epsfbox{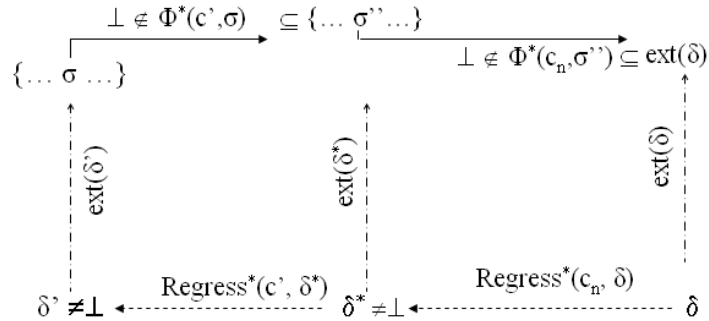}
    }
    \end{tabular}}
    \caption{Illustration of the proof of Lemma \ref{lem30b} - case 1.}
    \label{example-fig30b}
\end{figure}

        Let $c_n = a; p$ where $a$ is a sensing action, $p= case\ (\varphi_1 \rightarrow p_1
        \ldots \varphi_m \rightarrow p_m\ )$. Also, let us denote here $c_1; \ldots;
        c_{n-1}$ by $c'$. By Definition \ref{d3}, we have that

        \[\bot \neq \delta' = Regress^*(c,\delta) = Regress^*(c',Regress^*(c_n,\delta)).\]

        Denote $Regress^*(c_n,\delta) = \delta^*$. It follows from Lemma \ref{lem30} that $\delta^*$ is a
        p-state. Since $\delta' \neq \bot$, we also have that $\delta^* \neq \bot$.
        We first show that for every $\sigma'' \in ext(\delta^*)$,
        we have that $\bot \not \in \Phi^*(c_n,\sigma'')$ and $\Phi^*(c_n,\sigma'') \subseteq ext(\delta)$.

        Indeed, since
        $count(c_n) = 1 + \sum^m_{j=1} count(p_j) \leq count(c) \leq
        k+1$, we have that $count(p_i) \leq k$ for $i = 1,\ldots,m$. By Definition \ref{d3}:
        $$ \bot \neq \delta^* = Regress^*(c_n,\delta) = Regress(a,\{R(p_1,\delta), \ldots, R(p_m,\delta)\})$$

        Let's denote $R(p_i,\delta)$ by $\delta_i$ and $\Delta = \{\delta_1, \ldots,\delta_m\}$.
        We have that $\delta_i \models \varphi_i$ for $1 \leq i \leq m$, and $a$ is applicable in $\Delta$.

        From the proof of Proposition \ref{lem2}, we
        have $\bot \not \in \Phi(a,\sigma'')$ and

        \[\Phi(a,\sigma'') =
        \{{\sigma_1}'', \ldots, {\sigma_k}''\} \subseteq ext(\delta_1)
        \cup \ldots \cup ext(\delta_m)\]

        where for every ${\sigma_i}'' \in \Phi(a, \sigma'')$, there
        exists $1 \leq j \leq m$ such that ${\sigma_i}'' \in ext(\delta_j)$
        ($i=1.\ldots,k$). It is easy to see that $k \leq m$.
        Indeed, since $a$ is applicable in $\Delta$, by Definitions
        \ref{app-sensing} and \ref{sensed-Set} we have that $Sens_a$ is known
        in $\Delta$ and that $m = 2^{|S_a,\Delta|}$. By Definition \ref{reg-sensing} we have
        that $S_{a,\Delta}$ is the maximal set of sensing fluents that is unknown in
        $\delta^*$. Since $\sigma'' \in ext(\delta^*)$, i.e. $\delta^*.T \subseteq
        sigma''.T$ and $\delta^*.F \subseteq \sigma''.F$, by
        Definition \ref{def-tran} we have that $Sens_a \setminus \sigma'' \subseteq S_{a,\Delta}$.
        This implies that $k \leq m$. Using the Observation \ref{ob1}, item 2 we
        have that $\delta_j \models \varphi_i$ implies ${\sigma_i}'' \models
        \varphi_i$ ($i=1.\ldots,k$). As $\delta_i \models \varphi_i$
        for $1 \leq i \leq m$ and we can always arrange the order of
        elements of the set $\Phi(a,\sigma'')$, we can assume that ${\sigma_i}'' \models \varphi_i$
        ($i=1,\ldots,k$).

        From the definition of $\Phi^*$

        \[\Phi^*(c_n,\sigma'') =
        \bigcup_{\sigma' \in \Phi(a,\sigma'')} E(p,\sigma')
        = \Phi^*(p_1,{\sigma_1}'') \cup \ldots \cup
        \Phi^*(p_k,{\sigma_k}'').\]

        As $Regress^*(p_i,\delta).T \subseteq R(p_i,\delta).T$ and
        $Regress^*(p_i,\delta).F \subseteq R(p_i,\delta).F$,
        ${\sigma_i}'' \in ext(\delta_i)$ implies
        ${\sigma_i}'' \in ext(Regress^*(p_i,\delta))$.
        Using inductive hypothesis for $count(p_i) \leq
        k$, we have $\bot \not \in \Phi^*(p_i,{\sigma_i}'')$ and $\Phi^*(p_i,{\sigma_i}'') \subseteq
        ext(\delta)$ ($i=1.\ldots,k$). This means that $\bot \not \in \Phi^*(c_n,\sigma'')$ and $\Phi^*(c_n,\sigma'')
        \subseteq ext(\delta)$ [*].

        We have that
        \[\delta' = Regress^*(c,\delta) = Regress^*(c',\delta^*).\]
        Consider an arbitrary $\sigma \in ext(\delta')$. Since $count(c_n) \geq
        1$, we have that $count(c') \leq k$. Using the inductive hypothesis, we have
        that $\bot \not \in \Phi^*(c',\sigma)$ and $\Phi^*(c',\sigma)
        \subseteq ext(\delta^*)$.

        We will now continue with our proof. From the definition of $\Phi^*$, we have $\Phi^*(c,\sigma) =
        \bigcup_{\sigma' \in \Phi^*(c',\sigma)} \Phi^*(c_n,\sigma')$.
        Since $\Phi^*(c',\sigma) \subseteq ext(\delta^*)$, by using [*] we have that
        $\bot \not \in \Phi^*(c,\sigma)$ and $\Phi^*(c,\sigma) \subseteq ext(\delta)$.

        \item $c_n$ is a sequence of non-sensing actions:

        Let $c' = c_1; \ldots; c_{n-1}$. From Observation \ref{ob1},
        item 1, $c_{n-1}$ is a case plan.
        Since $count(c_n) = 0$, using case 1 above and Corollary
        \ref{cor1}, we can prove this second case.
    \end{enumerate}

\st From cases 1 and 2, the lemma is proved.
\end{itemize}
\end{proof}

\begin{figure}[h]
    \centerline{
    \begin{tabular}{c}
    {
    \epsfxsize=3in \epsfbox{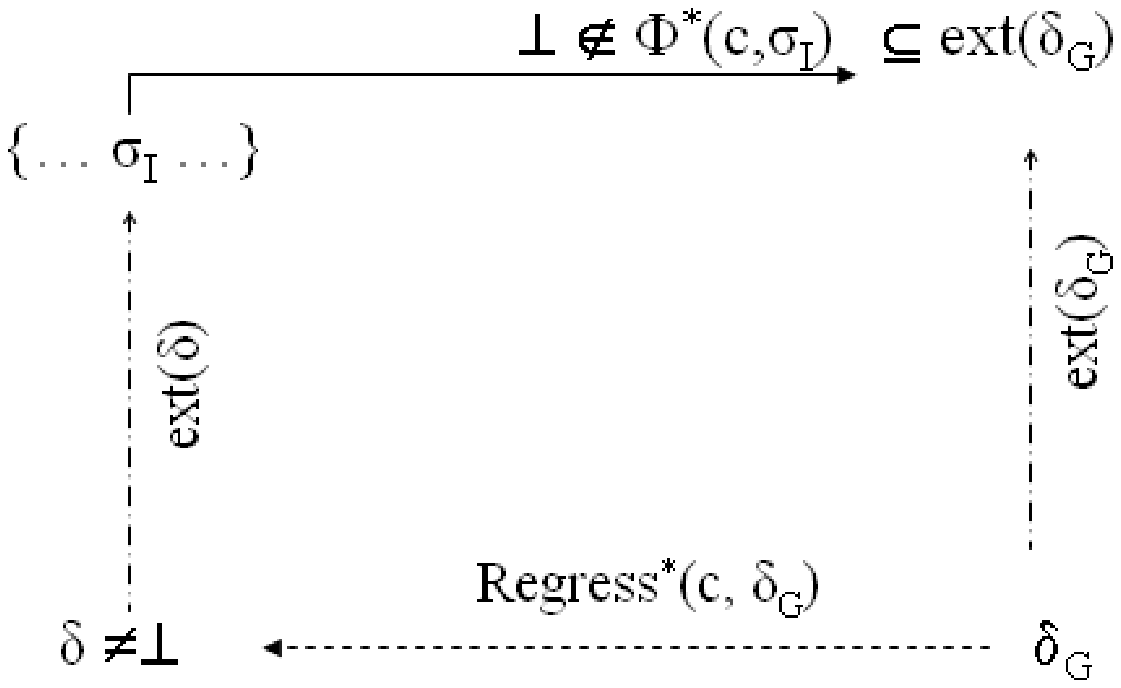}
    }
    \end{tabular}}
    \caption{Illustration for Theorem \ref{lem3}.}
    \label{example-fig1}
\end{figure}

\setcounter{section}{3}

\setcounter{theorem}{11}

{\theorem [Soundness of Regression] \label{lem3} Let  $P=\langle
A,O,I,G \rangle$ be a planning problem and $c$ be a regression
solution of $P$. Then, $\bot \not\in \Phi^*(c,\sigma_I)$ and
$\Phi^*(c,\sigma_I) \subseteq ext(\delta_G)$. }

\begin{proof} [Theorem \ref{lem3}] Let $\delta' =
Regress^*(c,\delta_G)$. Since $\delta' \neq \bot$ and $\sigma_I
\in ext(\delta')$ (from the definition of a regression solution),
the conclusion of the theorem follows immediately from Lemma
\ref{lem30b}. \end{proof}

\setcounter{section}{3}

\setcounter{theorem}{16}

\begin{lemma} \label{lemAdd1}
For every conditional plan $c$,

\begin{itemize}
\item [$\bullet$] $normalized(c)$ is a normalized conditional
plan;

\item [$\bullet$] for every a-state $\sigma$, $\Phi^*(c,\sigma) =
\Phi^*(normalized(c), \sigma)$.

\end{itemize}
\end{lemma}

\begin{proof} [Lemma \ref{lemAdd1}]
We prove by induction on $count(c)$. The base case is trivial
since $count(c)=0$ means that $c$ is a sequence of non-sensing
actions, which implies that $normalized(c) = c$ is a sequence of
non-sensing actions, which is a normalized plan. This also implies
that $\Phi^*(c,\sigma) = \Phi^*(normalized(c), \sigma)$ for every
a-state $\sigma$.

\st Assume that we have proved the lemma for $count(c) \le k$. Let
$c$ be a plan with $count(c) = k+1$. As we can write $c =
c_1;\ldots;c_n$, we have two cases:

\begin{itemize}
\item [$\bullet$]  $c_1$ is a case plan, $c_1 = a; case\
(\varphi_1 \rightarrow p_1
        \ldots \varphi_m \rightarrow p_m\ )$.
So, we have that $normalized(c) = a; case\ (\varphi_1 \rightarrow
normalized(p_1;c')
        \ldots \varphi_m \rightarrow normalized(p_m;c')\ )$
where $c' = c_2;\ldots;c_n$. Let $c''_i = p_i;c'$, we have that
$count(c''_i) \le k$. So, $normalized(p_i;c')$ is a normalized
plan. By construction of $normalized(c)$, we conclude that it is
indeed a normalized plan.

\st Let $\sigma$ be an arbitrary a-state.
We have that \\
$\Phi^*(c, \sigma) = \Phi^*(c', \Phi^*(c_1,\sigma)) $\\
\hspace*{0.48in}   $ = \bigcup_{\sigma'' \in \bigcup_{\sigma' \in
\Phi(a,\sigma)}
            E(case\ (\varphi_1 \rightarrow p_1)
        \ldots \varphi_m \rightarrow p_m\ ), \sigma')} \Phi^*(c', \sigma'')$ \\

\hspace*{0.48in}   $ = \bigcup_{\sigma' \in \Phi(a,\sigma)}
            E(case\ (\varphi_1 \rightarrow (p_1;c')
        \ldots \varphi_m \rightarrow (p_m;c')\ ), \sigma')$ \\
\hspace*{0.48in}   $= \Phi^*(normalized(c), \sigma)$.

\st Note that the last equation follows from the inductive
hypothesis that $\Phi^*(p_i;c', \sigma) =
\Phi^*(normalized(p_i;c'), \sigma)$ for every a-state $\sigma'$.

\item [$\bullet$] $c_1$ is a sequence of non-sensing actions.
Then, $c_2$ is a case plan. Similar arguments as in the previous
case allow us to conclude that $normalized(c')$, where $c' =
c_2;\ldots;c_n$,  is a normalized plan and is a case plan.
Furthermore, for every a-state $\sigma$, $\Phi^*(c',\sigma) =
\Phi^*(normalized(c'), \sigma)$. Thus, $normalized(c) = c_1;
normalized(c')$ is a normalized plan and $\Phi^*(c,\sigma) =
\Phi^*(c_1;c', \sigma) = \Phi^*(c_1;normalized(c), \sigma) =
\Phi^*(normalized(c), \sigma)$. 
\end{itemize}
\end{proof}

{\lemma \label{factorable} Let $\chi = \{\varphi_1, \ldots,
\varphi_n\}$ be a non-empty set of consistent conjunctions of
literals. If $\chi$ is factorable, then there exists a unique
non-empty set of fluents $S$ such that $\chi$ spans over $S$.}

\begin{proof} [Lemma \ref{factorable}] Since $\chi$ is factorable,
there exists a non-empty set of fluents $S$ such that $\chi$ spans
over $S$, i.e. there exists $\varphi$ such that $\varphi_i =
\varphi \wedge \psi_i$ where $\psi_i \in BIN(S)$ for $i=1,
\ldots,n$. Assume that $S$ is not unique. As a result, there
exists a non-empty set $S' \neq S$ such that $\chi$ spans over
$S'$, i.e. there exists $\varphi'$ such that $\varphi_i = \varphi'
\wedge \psi'_i$ where $\psi'_i \in BIN(S')$ for $i=1, \ldots,n$.

\st Consider $f \in S \setminus S'$. For every $1 \leq i \leq n$,
we have that $\varphi_i = \varphi' \wedge \psi'_i$. Since $f \not
\in S'$ and $\varphi_i$ is consistent ($1 \leq i \leq n$), $f$
must occur either positively or negatively in $\varphi'$. This means
that $f$ occurs either positively or negatively in all $\varphi_i$ for
$1 \leq i \leq n$.

\st Consider the case that $f$ occurs positively in all
$\varphi_i$ for $1 \leq i \leq n$ [*]. Since $f \in S$, there
exists a binary representation $\psi_j \in BIN(S)$ ($1 \leq j \leq
n$) such that $f$ appears negatively in $\psi_j$ i.e. $f$ appears
negatively in $\varphi_j$. This contradicts with [*]. Similarly we
can show a contradiction in the case that $f$ occurs negatively in
all $\varphi_i$ for $1 \leq i \leq n$. We conclude that $S$ is
unique. \end{proof}

\setcounter{section}{8}

\setcounter{theorem}{5}

{\lemma \label{lem311} Let $\sigma$ be an a-state, $\delta$ be a
p-state, and $c=a_1; \ldots;a_n$ ($n \geq 1$) be a sequence of
non-sensing actions. Assume that $c$ is regressable with respect
to $(\sigma,\delta)$. Then, $Regress^*(a_n,\delta) = \delta^*$,
$\delta^* \neq \bot$, and $a_1,\ldots,a_{n-1}$ is regressable with
respect to $(\sigma,\delta^*)$. }

\begin{proof} [Lemma \ref{lem311}]

We prove by induction on $|c|$, the number of actions in $c$.
\begin{itemize}
    \item [$\bullet$] Base Case: $|n|=1$. Similar to the inductive step,
we can show that $a_1$ is applicable in $\delta$. Let
$\delta^*=Regress(a_1,\delta)$ and $\Phi(a_1,\sigma) =
\{\sigma'\}$. We have that, $\delta^*.T = \delta.T \setminus Add_a
\cup Pre_a^+$ and $\sigma'.T = \sigma.T \setminus Del_a \cup
Add_a$. Using the facts $\sigma' \in ext(\delta)$ and $Add_a \cap
Del_a = \emptyset$ and the above equations, we can show that
$\delta^*.T \subseteq \sigma.T$. Similarly, $\delta^*.F \subseteq
\sigma.F$. Since $[\ ]$ is not redundant with respect to
$(\sigma,\delta^*)$, we have that $[\ ]$ is a plan that is
regressable with respect to $(\sigma,\delta^*)$.

    \item [$\bullet$] Inductive Step: Assume that we have proved the lemma
    for $0 < n \le k$. We need to prove the lemma for $n=k+1$.

    Let $\Phi^*(a_1; \ldots; a_k,\sigma) = \{\sigma_k\}$, we have
    that
    \[\Phi^*(c,\sigma) = \Phi(a_{k+1},\sigma_k) = \{\sigma'\} \subseteq
    ext(\delta).\]

    We will prove that (1) $a_{k+1}$ is applicable in $\delta$,
    (2) $\sigma_k \in ext(\delta^*)$ where $\delta^* = Regress(a_{k+1},\delta)$ and
    $\delta^* \neq \bot$, and (3) $c'=a_1, \ldots, a_k$ is
    regressable with respect to $(\sigma,\delta^*)$.

    \begin{itemize}
        \item Proof of (1):
    We first show that
    $Add_{a_{k+1}} \cap \delta.T \neq \emptyset$ or
    $Del_{a_{k+1}} \cap \delta.F \neq \emptyset$.
    Assume the contrary, $Add_{a_{k+1}} \cap \delta.T = \emptyset$ and
    $Del_{a_{k+1}} \cap \delta.F = \emptyset$.
    By Definition \ref{def-tran}, we have that
    $\sigma'.T = \sigma_k.T \setminus Del_{a_{k+1}} \cup Add_{a_{k+1}}$
    and
    $\sigma'.F = \sigma_k.F \setminus Add_{a_{k+1}} \cup
    Del_{a_{k+1}}$.
    Since $\sigma' \in ext(\delta)$, we have $\delta.T \subseteq
    \sigma'.T$. By our assumption, $Add_{a_{k+1}} \cap \delta.T = \emptyset$,
    we must have that $\delta.T = \delta.T \setminus Add_{a_{k+1}} \subseteq \sigma'.T \setminus
    Add_{a_{k+1}}$. Because for arbitrary sets $X,Y$,
    $(X \cup Y) \setminus Y = X \setminus (X \cap Y)$, we have that
    \[\sigma'.T \setminus Add_{a_{k+1}} = ((\sigma_k.T \setminus Del_{a_{k+1}}) \cup Add_{a_{k+1}})
    \setminus Add_{a_{k+1}} =\] \[(\sigma_k.T \setminus Del_{a_{k+1}}) \setminus ((\sigma_k.T \setminus Del_{a_{k+1}}) \cap Add_{a_{k+1}}) \subseteq \sigma_k.T ,\]
    i.e. $ \delta.T \subseteq \sigma_k.T \setminus Del_{a_{k+1}}$.
    This shows that $\delta.T \subseteq \sigma_k.T$. Similarly, we can show that $\delta.F \subseteq
    \sigma_k.F$. We conclude that $\sigma_k \in ext(\delta)$, i.e.
    $c$ is redundant with respect to $(\sigma,\delta)$.
    This is a contradiction. Therefore, $Add_{a_{k+1}} \cap \delta.T \neq \emptyset$ or
    $Del_{a_{k+1}} \cap \delta.F \neq \emptyset$ (i).

    Secondly, as $\sigma' \in
    ext(\delta)$, we have $\delta.T \subseteq \sigma'.T$ and $\delta.F
    \subseteq \sigma'.F$. As $a_{k+1}$ is executable in $\sigma_k$, we have
    $Add_{a_{k+1}} \cap \sigma'.F = \emptyset$ and $Del_{a_{k+1}} \cap
    \sigma'.T = \emptyset$. This concludes that $Add_{a_{k+1}} \cap \delta.F =
    \emptyset$ and $Del_{a_{k+1}} \cap \delta.T = \emptyset$ (ii).

    Thirdly,
    assume that there exists $f \in Pre^+_{a_{k+1}} \cap
    \delta.F$ and $f \not \in Del_{a_{k+1}}$. By Definition
    \ref{def-tran}, it's easy to see
    that $f \in \sigma'.T$ and $f \in \sigma'.F$. This is a
    contradiction, therefore $Pre^+_{a_{k+1}} \cap \delta.F \subseteq
    Del_{a_{k+1}}$. Similarly, we can show that $Pre^-_{a_{k+1}} \cap
    \delta.T \subseteq
    Add_{a_{k+1}}$ (iii). From (i), (ii), and (iii) we conclude
    that $a_{k+1}$ is applicable in $\delta$.

    \item Proof of (2):
    Because $a_{k+1}$ is applicable in $\delta$, we have that
    $Regress(a_{k+1},\delta) = \delta^*$ and $\delta^* \neq \bot$.
    We will show that $\sigma_k \in ext(\delta^*)$:

    Indeed, as $\sigma' \in ext(\delta)$, by Definition \ref{def-tran} we
    have
    \[\delta.T \subseteq \sigma'.T = \sigma_k.T \setminus Del_{a_{k+1}}
    \cup
    Add_{a_{k+1}}\] and
    \[\delta.F \subseteq \sigma'.F = \sigma_k.F \setminus Add_{a_{k+1}}
    \cup
    Del_{a_{k+1}}.\]

    By Definition \ref{reg-nonsensing} we have
    \[\delta^*.T = \delta.T \setminus Add_{a_{k+1}} \cup
    Pre^+_{a_{k+1}}\] and
    \[\delta^*.F = \delta.F \setminus Del_{a_{k+1}} \cup
    Pre^-_{a_{k+1}}.\]
    Since $a_{k+1}$ is executable in $\sigma_k$, we have that
    $Pre^+_{a_{k+1}} \subseteq \sigma_k.T$
    and $Pre^-_{a_{k+1}} \subseteq \sigma_k.F$. Therefore, to prove
    that $\delta^*.T = \delta.T \setminus Add_{a_{k+1}} \cup
    Pre^+_{a_{k+1}} \subseteq \sigma_k.T$, we only need to show that
    $\delta.T \setminus Add_{a_{k+1}} \subseteq \sigma_k.T$. As
    $\delta.T \subseteq \sigma_k.T \setminus Del_{a_{k+1}} \cup
    Add_{a_{k+1}}$, we have
    \[\delta.T \setminus Add_{a_{k+1}} \subseteq ((\sigma_k.T \setminus
    Del_{a_{k+1}}) \cup Add_{a_{k+1}}) \setminus Add_{a_{k+1}}.\]
    From the proof of item (1) above, we have that $((\sigma_k.T \setminus
    Del_{a_{k+1}}) \cup Add_{a_{k+1}}) \setminus Add_{a_{k+1}} \subseteq
    \sigma_k.T$. This concludes that $\delta.T \setminus Add_{a_{k+1}} \subseteq
    \sigma_k.T$. Similarly, we can show that $\delta.F \setminus
    Del_{a_{k+1}} \subseteq
    \sigma_k.F$, i.e., $\sigma_k \in ext(\delta^*)$ or $\{\sigma_k\}
    \subseteq ext(\delta^*)$.

    \item Proof of (3):
    Suppose that $c'$ is redundant with respect to $(\sigma,\delta^*)$.
    By Definition \ref{redundancy}, there exists a subplan $c''$ of $c$
    such that $\bot \not \in \Phi^*(c'', \sigma)$ and
    $\Phi^*(c'', \sigma) = \{\sigma''\} \subseteq ext(\delta^*)$.
    By Lemma \ref{lem1}, we have that $\bot \not \in
    \Phi(a_{k+1},\sigma'') \subseteq
    ext(\delta)$. Since \[\Phi^*(c''; a_{k+1}, \sigma) =
    \Phi(a_{k+1}, \sigma'') \subseteq ext(\delta),\]
    we have that $c$ is redundant with respect to
    $(\sigma,\delta)$. This
    contradicts with the assumption that $c$ is not redundant
    with respect to $(\sigma,\delta)$. Since $c'$ has no case
    plan, this concludes that $c'=a_1, \ldots, a_k$ is not
    redundant with respect to $(\sigma,\delta^*)$. Since
    $\bot \not \in \Phi^*(a_1; \ldots; a_k,\sigma) = \{\sigma_k\} \subseteq ext(\delta^*)$
    we have that $c'$ is regressable with respect to $(\sigma,\delta^*)$.

\end{itemize}
\end{itemize}
\end{proof}

{\lemma \label{lem31} Let $\sigma$ be an a-state and $\delta$ be a
p-state. Let $c=a_1; \ldots;a_n$ be a sequence of non-sensing
actions that is regressable with respect to $(\sigma,\delta)$.
Then, it holds that $Regress^*(c,\delta) = \delta^*$, $\delta^*
\neq \bot$, and $\sigma \in ext(\delta^*)$. }

\begin{proof} [Lemma \ref{lem31}]


We prove by induction on $|c|$, the number of actions in $c$.
\begin{itemize}
    \item [$\bullet$] Base Case: $|c|=0$. Then $c$ is an empty sequence of
    non-sensing actions. The base case follows from Definition \ref{d3}
    (with $\delta^* = \delta$. Note that $[\ ]$ is not a redundant
action).

    \item [$\bullet$] Inductive Step: Assume that we have proved the
    lemma for $|c| = k \geq 0$. We need to prove the lemma for
    $|c|=k+1$.
    It follows from Lemma \ref{lem311}
    that $\delta' = Regress(a_{k+1},\delta)$,
    $\delta' \ne \bot$, and
    $c' = a_1;\ldots;a_{k}$ is a plan that is regressable
    with respect to $(\sigma,\delta')$.
    By inductive hypothesis,  we have that
    $Regress^*(c',\delta') =\delta^* \ne \bot$
    and $\sigma \in ext(\delta^*)$.
    The inductive step follows from this and the fact
    $Regress^*(c,\delta) = Regress^*(c',Regress(a_{k+1},\delta))$.
\end{itemize}
\end{proof}

\COMMENT

{\lemma \label{lem31a} Let $\sigma$ be an a-state, $\delta$ be a
p-state, and $c$ be a conditional plan of the form $c=c_1; \ldots;
c_{n-1}; c_n$ where $c_1, \ldots, c_{n-1}$ are conditional plans,
$c_n=a_1; \ldots; a_k; a_{k+1}$ is a sequence of non-sensing
actions. Let $c'=c_1; \ldots; c_{n-1};a_1; \ldots;a_k$. If $c$ is
not redundant with respect to $(\sigma, \delta)$ then
$Regress^*(a_{k+1}, \delta) = \delta' \neq \bot$ and $c'$ is not
redundant with respect to $(\sigma,\delta')$.}

\st \proof{[Lemma \ref{lem31a}] We will use Lemma \ref{lem311}
in our proof.

\st By Definition \ref{def-tran}, we have that \[\bot \not \in
\Phi^*(c,\sigma) = \bigcup_{\sigma_k \in \Phi^*(c',\sigma)}
\Phi(a_{k+1}, \sigma_k) \subseteq ext(\delta).\] This shows that
$\bot \not \in \Phi^*(c',\sigma)$ and for every $\sigma_k \in
\Phi^*(c',\sigma)$, then $\Phi(a_{k+1}, \sigma_k) = \{\sigma'\}
\subseteq ext(\delta)$.

\st We first need to show that $a_{k+1}$ is applicable in
$\delta$, i.e. we have to prove that (i) $Add_{a_{k+1}} \cap
\delta.T = \emptyset$ or $Del_{a_{k+1}} \cap \delta.F =
\emptyset$; (ii) $Add_{a_{k+1}} \cap \delta.F = \emptyset$ and
$Del_{a_{k+1}} \cap \delta.T = \emptyset$; and (iii)
$Pre^+_{a_{k+1}} \cap \delta.F \subseteq Del_{a_{k+1}}$ and
$Pre^-_{a_{k+1}} \cap \delta.T \subseteq Add_{a_{k+1}}$. In
proving (i), if we assume that $Add_{a_{k+1}} \cap \delta.T =
\emptyset$ and $Del_{a_{k+1}} \cap \delta.F = \emptyset$ then,
using similar arguments in case (1), item ``Inductive Step'' of
Lemma \ref{lem311}, we conclude that $\sigma_k \in ext(\delta)$.
This shows that, for every $\sigma_k \in \Phi^*(c',\sigma)$ then
$\sigma_k \in ext(\delta)$, i.e. $c$ contains $a_{k+1}$ as a
redundant action with respect to $(\sigma,\delta)$. This is a
contradiction. The proof of (ii) and (iii) also follows directly
from arguments in case (1), item ``Inductive Step'' of Lemma
\ref{lem311}. Therefore, $a_{k+1}$ is applicable in $\delta$ [*].

\st Using similar arguments in case (2), item ``Inductive Step''
of Lemma \ref{lem311}, we can show that $\sigma_k \in
ext(\delta')$ where $Regress(a_{k+1}, \delta) = \delta' \neq
\bot$. Therefore $\bot \not \in \Phi^*(c',\sigma) \subseteq
ext(\delta')$ [**].

\st We now show that $c'$ is not redundant with respect to
$(\sigma,\delta')$. Assume the contrary, we then have a subplan
$c''$ of $c'$ such that $\bot \not \in \Phi^*(c'',\sigma)
\subseteq ext(\delta')$. We have that
\[\Phi^*(c'';a_{k+1}, \sigma) = \bigcup_{\sigma' \in
\Phi^*(c'',\sigma)} \Phi(a_{k+1}, \sigma').\] Since
$\Phi^*(c'',\sigma) \subseteq ext(\delta')$, from Lemma \ref{lem1}
we have that $\Phi^*(c'';a_{k+1}, \sigma) \subseteq ext(\delta)$.
Since $c'';a_{k+1}$ is a subplan of $c$, therefore $c$ is
redundant with respect to $(\sigma, \delta)$. This is a
contradiction [***]. From [*], [**], [***], the lemma is proved.
\hfill $\Box$ }

\ENDCOMMENT

\COMMENT {\corollary \label{lem31aa} Let $\sigma$ be an a-state,
$\delta$ be a p-state, and $c$ be a sequence of non-sensing
actions. If (i) $\bot \not \in \Phi^*(c,\sigma) \subseteq
ext(\delta)$; and (ii) $Regress^*(c, \delta) = \delta' \neq \bot$
then $\sigma \subseteq ext(\delta')$.}

\st \proof{[corollary \ref{lem31a}] We prove by induction on
$|c|$, the number of actions in $c$.

\begin{itemize}
    \item [$\bullet$] Base Case: $|c|=1$.
    \ref{lem31}.
    \item [$\bullet$] Inductive Step: Assume that we have proved the lemma
    for $count(c) \leq k$. We need to prove the lemma for
    $count(c)=k+1$.
\end{itemize} }

\ENDCOMMENT

{\lemma \label{lem31ab} Let $\sigma$ be an a-state, $a$ be a
sensing action which is executable in $\sigma$.
Let $S_a = Sense_a \setminus \sigma$.
Then, we have that (1) $\Phi(a,\sigma) = \{\sigma_1,
\ldots, \sigma_m\}$ where $m=2^{|S_a|}$, (2) $a$ is strongly
applicable in $\Delta = \{\delta_1, \ldots, \delta_m\}$ where
$\delta_i = [\sigma_i.T, \sigma_i.F]$, $i=1,\ldots,m$, and (3)
$Regress(a, \Delta) = [\sigma.T, \sigma.F]$.}

\begin{proof} [Lemma \ref{lem31ab}]:

\st Proof of (1): From Definition \ref{def-tran}, we have that
\[\bot \not \in \Phi(a,\sigma) = \{\sigma' | Sens_a \setminus \sigma = \sigma' \setminus \sigma\}.\]
We have that, for every $\sigma' \in \Phi(a,\sigma)$ then $\sigma'
\setminus \sigma = (\sigma'.T \setminus \sigma.T) \cup (\sigma'.F
\setminus \sigma.F)$. Denote $\sigma'.T \setminus \sigma.T$ by $P$
and $\sigma'.F \setminus \sigma.F$ by $Q$, we have that $(P,Q)$ is
a partition of $S_a$. Since there are $2^{|S_a|}$ partitions of
$S_a$, we have that $m \leq 2^{|S_a|}$. Furthermore, for a
partition $(P,Q)$ of $S_a$ we have that there exists an a-state
$\sigma' = \langle P \cup \sigma.T, Q \cup \sigma.F \rangle \in
\Phi(a,\sigma)$ because $\sigma' \setminus \sigma = P \cup Q$.
Therefore $2^{|S_a|} \leq m$. We conclude that $m = 2^{|S_a|}$.

\st Proof of (2): We first show that $\Delta$ is proper with
respect to $S_a$, i.e. $S_a$ is a sensed set of $\Delta$ with
respect to $a$. Indeed, by Definition \ref{def-tran} and the proof
of (1) above, we have that the conditions \emph{(i)}-\emph{(iii)}
of Definition \ref{sensed-Set} are satisfied. The condition
\emph{(iv)} of Definition \ref{sensed-Set}  is satisfied because
we have that $\delta_i.T \setminus S_a = \sigma_i.T \setminus S_a
= \sigma.T$ and $\delta_i.F \setminus S_a = \sigma_i.F \setminus
S_a = \sigma.F$ ($1 \leq i \leq m$). Therefore, we conclude that
$\Delta$ is proper with respect to $S_a$.

\st Since $a$ is an action that is executable in $\sigma$ we have
that
$(Pre^+_a \cup Pre^-_a) \cap Sens_a = \emptyset$ and
$Pre^+_a \cap \sigma.F = \emptyset$, $Pre^-_a \cap \sigma.T =
\emptyset$, therefore $Pre^+_a \cap \delta_i.F = \emptyset$,
$Pre^-_a \cap \delta_i.T = \emptyset$ ($1 \leq i \leq m$). By
Definition \ref{app-sensing-strong}, we conclude that $a$ is
strongly applicable in $\Delta$.

\st Proof of (3): Since $a$ is executable in $\sigma$, we have
that $Pre^+_a \subseteq \sigma.T$ and $Pre^-_a \subseteq
\sigma.F$. From the proof of (2), $\delta_i.T \setminus S_a =
\sigma.T$ and $\delta_i.F \setminus S_a = \sigma.F$ ($1 \leq i
\leq m$). The proof follows from Definition \ref{reg-sensing}
where we let $S_a = S_{a,\Delta}$. \end{proof}

{\lemma \label{lem31b-son} Let $\sigma$ be an a-state, $\delta$ be a
p-state, and $c = \alpha; c'$ is a normalized conditional plan
where $\alpha$ is a non-empty sequence of sensing actions
and $c' = a; case(\varphi_1 \rightarrow p_1,\ldots,
          \varphi_m \rightarrow p_m)$.
If $c$ is regressable with respect to
$(\sigma,\delta)$. Then,

\begin{itemize}
\item [$\bullet$] $\Phi^*(\alpha, \sigma) = \{\sigma_1\}$ and
$\sigma_1 \ne \bot$;

\item [$\bullet$] $m = 2^{|S_a|}$ where $S_a = Sens_a \setminus
\sigma_1$;

\item [$\bullet$] $\{\varphi_1,\ldots,\varphi_m\}$ spans over
$S_a$;

\item [$\bullet$] For each $i$, $1 \le i \le m$, there exists a
unique a-state $\sigma' \in \Phi(a, \sigma_1)$ such that $p_i$ is
regressable with respect to $(\sigma',\delta)$.
\end{itemize}
}

\begin{proof} [Lemma \ref{lem31b-son}]:

\begin{itemize}
\item [$\bullet$] By Definition \ref{def-extran}, we have that
\[
\Phi^*(c, \sigma) = \bigcup_{\sigma' \in  \Phi^*(\alpha, \sigma)} \Phi^*(c', \sigma').
\]
Since $c$ is regressable with respect to $(\sigma,\delta)$ we
have that $\bot\not\in \Phi^*(c, \sigma)$.
This implies that $\bot \not\in \Phi^*(\alpha, \sigma)$.
Furthermore, because $\alpha$ is a sequence of non-sensing actions,
we conclude that $\Phi^*(\alpha, \sigma)$ is a singleton,
i.e., $\Phi^*(\alpha, \sigma) = \{\sigma_1\}$ for some a-state $\sigma_1$.
From $\bot \not\in \Phi^*(\alpha, \sigma)$, we have that
$\sigma_1 \ne \bot$.

\item [$\bullet$] By definition of $S_a$ we conclude that $S_a$ is
the set of fluents that belong to $Sens_a$ which are unknown in
$\sigma_1$. By Definition \ref{def-tran} we conclude that $\Phi(a,
\sigma_1)$ consists of $2^{|S_a|}$ elements where for each
$\sigma' \in \Phi(a, \sigma_1)$, $\sigma' \setminus \sigma_1 =
S_a$. Because
\[
\bot \not\in \Phi^*(c, \sigma) = \bigcup_{\sigma' \in  \Phi(a, \sigma_1)}
E(case(\varphi_1 \rightarrow p_1,\ldots,
          \varphi_m \rightarrow p_m), \sigma')
\]
we conclude that for each $\sigma' \in \Phi(a, \sigma_1)$
there exists one $j$, $1 \le j \le m$, such that
$\varphi_j$ is satisfied in $\sigma'$. Since
$\varphi$'s are mutual exclusive we conclude that for each $j$,
$1 \le j \le m$,
there exists at most one $\sigma' \in \Phi(a, \sigma_1)$
such that $\varphi_j$ is satisfied in $\sigma'$.
This implies that $m = 2^{|S_a|}$.

\item [$\bullet$] Since $c$ is regressable with respect to
$(\sigma,\delta)$ we have that $a; (case(\varphi_1 \rightarrow
p_1,\ldots, \varphi_m \rightarrow p_m)$ is possibly regressable.
This implies that $\{\varphi_1,\ldots,\varphi_m\}$ spans over a
set of fluents $S \subseteq Sens_a$ and there exists a $\varphi$
such that for every $i$, $\varphi_i = \psi_i \wedge \varphi$ where
$\psi_i \in BIN(S)$ and $S \cap (\varphi^+ \cup \varphi^-) =
\emptyset$. From Lemma \ref{factorable} we know that $S$ is
unique. We will show now that $S = S_a$. Assume the contrary, $S
\ne S_a$. We consider two cases:

\begin{itemize}
\item $S \setminus S_a \ne \emptyset$. Consider a fluent $f \in S
\setminus S_a$. Because $\{\varphi_1,\ldots,\varphi_m\}$ spans
over $S$, there exists some $i$ such that $f$ occurs positively in
$\varphi_i$. From the proof of the previous item and the fact that
$f \not\in S_a$, we conclude that $f$ must be true in $\sigma_1$
(otherwise, we have that the subplan $c'$ of $c$, obtained by
removing the branch $\varphi_i \rightarrow p_i$, satisfies $\bot
\not\in \Phi^*(c',\sigma) \subseteq ext(\delta)$, which implies
that $c$ is redundant with respect to $(\sigma,\delta)$).
Similarly, there exists some $j$ such that $f$ occurs negatively
in $\varphi_j$, and hence, $f$ must be false in $\sigma_1$. This
is a contradiction. Thus, this case cannot happen.

\item $S_a \setminus S \ne \emptyset$. Consider a fluent $f \in
S_a \setminus S$. Again, from the fact that $c$ is regressable
with respect to $(\sigma,\delta)$, we conclude that $f$ occurs
either positively or negatively in $\varphi_i$. Because $f \not\in
S$, we have that $f$ occurs in $\varphi$, and hence, $f$ occurs
positively or negatively in all $\varphi_i$. In other words, $f$
is true or false in every $\sigma' \in \Phi(a,\sigma_1)$. Thus,
$f$ is true or false in $\sigma_1$. This contradicts the fact that
$f \in S_a = Sens_a \setminus \sigma_1$. Thus, this case cannot
happen too.
\end{itemize}

The above two cases imply that $S_a = S$. This means that
$\{\varphi_1,\ldots,\varphi_m\}$ spans over $S_a$.

\item [$\bullet$] Consider an arbitrary $i$, $1 \le i \le m$. From
the proof of the second item, we know that there exists a unique
$\sigma' \in \Phi(a,\sigma_1)$ such that $\varphi_i$ is satisfied
by $\sigma'$. We will show now that $p_i$ is regressable with
respect to $(\sigma',\delta)$. From the fact that $c$ is
regressable, we conclude that every case plan in $p_i$ is possibly
regressable. Furthermore, because $\Phi^*(p_i, \sigma') \subseteq
\Phi^*(c, \sigma)$, we have that $\bot \not\in \Phi^*(p_i,
\sigma') \subseteq ext(\delta)$. Thus, to complete the proof, we
need to show that $p_i$ is not redundant with respect to
$(\sigma', \delta)$. Assume the contrary, there exists a subplan
$p'$ of $p_i$ such that $\bot \not\in \Phi^*(p', \sigma')
\subseteq ext(\delta)$. This implies that the subplan $c'$ of $c$,
obtained by replacing $p_i$ with $p'$, will satisfy that $\bot
\not\in \Phi^*(c', \sigma) \subseteq ext(\delta)$, i.e., $c$ is
redundant with respect to $(\sigma, \delta)$. This contradicts the
condition of the lemma, i.e., our assumption is incorrect. Thus,
$p_i$ is not redundant with respect to $(\sigma', \delta)$, and
hence, $p_i$ is regressable with respect to $(\sigma', \delta)$.
\end{itemize}
\end{proof}

\setcounter{section}{3}

\setcounter{theorem}{20}

{
\lemma \label{lem31b} Let $\sigma$ be an a-state, $\delta$ be a
p-state, and $c$ is a normalized conditional plan that is
regressable with respect to
$(\sigma,\delta)$. Then, $Regress^*(c,\delta) = \delta'$, $\delta'
\neq \bot$, and $\sigma \in ext(\delta')$.
}

\begin{proof} [Lemma \ref{lem31b}]
\begin{figure}[h]
    \centerline{
    \begin{tabular}{c}
    {
    \epsfxsize=3.5in \epsfbox{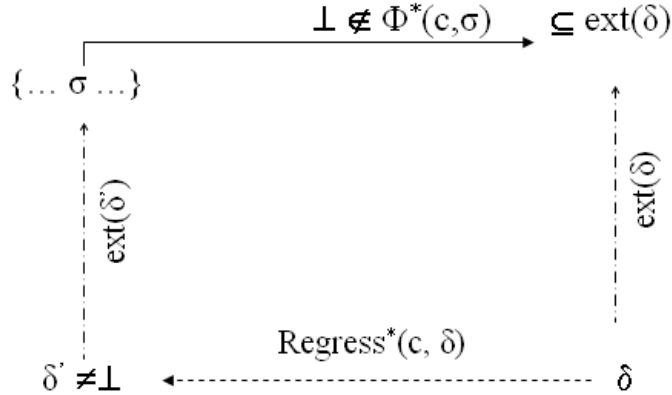}
    }
    \end{tabular}}
    \caption{Illustration of Lemma \ref{lem31b}.}
    \label{example-fig31b}
\end{figure}
We will prove by induction on $count(c)$,
the number of case plans in $c$.

\begin{itemize}

\item [$\bullet$] Base Case: $count(c)=0$. Then $c$ is a sequence
of non-sensing actions. The base case follows from Lemma
\ref{lem31}.

\item [$\bullet$] Inductive Step:
    Assume that we have proved the lemma
    for $count(c) \leq k$. We need to prove the lemma for
    $count(c)=k+1$.
Since $c$ is a normalized conditional plan, by Definition \ref{normalizedPlan},
we have that $c = \alpha; c'$ where $\alpha$ is a sequence of
non-sensing actions and $c' = a; p$ and $p= case\ (\varphi_1 \rightarrow p_1
        \ldots \varphi_m \rightarrow p_m\ )$.
Because $\alpha$ is a sequence of non-sensing actions
we have that $\Phi^*(\alpha, \sigma)$ is a singleton. Let
$\Phi^*(\alpha, \sigma) = \{\sigma_1\}$.

\st Let $S_a = Sens_a \setminus \sigma_1$. Since $c$ is not
redundant with respect to $(\sigma,\delta)$ we conclude that $S_a
\ne \emptyset$.

\st It follows from the fact that  $c$ is regressable with respect
to $(\sigma,\delta)$ and Lemma \ref{lem31b-son} that $\{\varphi_1,
\ldots, \varphi_n\}$ spans over $S_a$ and for every $i$, $1 \le i
\le m$, there exists a unique $\sigma' \in \Phi(a, \sigma_1)$ such
that $p_i$ is regressable with respect to $(\sigma',\delta)$. By
inductive hypothesis for $p_i$, we conclude that $Regress^*(p_i,
\delta) = \delta_i \ne \bot$ and $\sigma' \in ext(\delta_i)$.
Because $\varphi_i$ is satisfied by $\sigma'$ we have that
$R(p_i,\delta) = [\delta_i.T \cup \varphi_i^+, \delta_i.F \cup
\varphi_i^-]$ is consistent and hence $R(p_i,\delta) \ne \bot$.
This also implies that $\sigma' \in ext(R(p_i,\delta))$ and
$R(p_i,\delta) \ne R(p_j,\delta)$ for $i \ne j$.

\st
Let $\Delta = \{R(p_i,\delta) \mid i=1,\ldots,m\}$. We will show next
that $a$ is applicable in $\Delta$. Consider $\Delta' = \Phi(a, \sigma_1)$,
we have that for each $i$, $1 \le i \le m$, there exists
one $\sigma' \in \Delta'$ and $\sigma' \in ext(R(p_i,\delta))$.
It follows from Lemma \ref{lem31ab} that $a$ is strong applicable in $\Delta'$.
Thus, $a$ is applicable in $\Delta$.

\st
By definition of $Regress$, we have that

\[
Regress(a,\Delta) =
[\bigcup_{i=1}^m R(p_i,\delta).T \setminus S_a \cup Pre^+_a ,
      \bigcup_{i=1}^m R(p_i,\delta).F \setminus S_a \cup Pre^-_a
       ] = \delta^* \ne \bot.
\]

Since $a$ is executable in $\sigma_1$, from Lemma \ref{lem31ab}, and the fact that
for each $\sigma' \in \Phi(a,\sigma_1)$ there exists an $i$ such that
$\sigma' \in ext(R(p_i,\delta))$,
we can conclude $\sigma_1 \in ext(\delta^*)$.

\st To continue our proof, we will now show that $q = \alpha$ is
not redundant with respect to $(\sigma, \delta^*)$. Assume the
contrary, there exists a subplan $q'$ of $q$ such that $\bot
\not\in \Phi^*(q, \sigma) \subseteq ext(\delta^*)$. This, together
with the fact that $Regress^*(c', \delta) = \delta^*$ and the
soundness theorem \ref{lem3} implies that $\bot \not\in
\Phi^*(c'', \sigma) \subseteq ext(\delta)$ for $c'' = q';c'$,
i.e., $c$ is redundant with respect to $(\sigma,\delta)$. This
contradicts the assumption of the lemma, i.e., we have proved that
$q$ is not redundant with respect to $(\sigma, \delta^*)$.

\st Applying the inductive hypothesis for the plan $q$ and
$(\sigma,\delta^*)$, we have that $Regress^*(q, \delta^*) =
\delta' \ne \bot$ and $\sigma \in ext(\delta')$. The inductive
hypothesis is proved because $Regress^*(c, \delta) = Regress^*(q,
\delta^*)$.  
\end{itemize}
\end{proof}

\setcounter{section}{8}

\setcounter{theorem}{9}

\st {\lemma \label{lemAdd2} Let $\sigma$ be an a-state, $\delta$
be a p-state, and $c$ be a sequence of non-sensing actions such
that $\bot \not \in \Phi^*(c,\sigma)$ and $\Phi^*(c,\sigma)
\subseteq ext(\delta)$. Then, there exists a subplan $c'$ of $c$
that is not redundant with respect to $(\sigma,\delta)$ and $c'$
is equivalent to $c$ with respect to $(\sigma,\delta)$.}

\begin{proof} [Lemma \ref{lemAdd2}] Notice that the length of $c$ is finite. Consider two cases:
\begin{itemize}
    \item [$\bullet$] Case (i): $c$ is not redundant with respect to $(\sigma,\delta)$.

    It's easy to see that $c'=c$ satisfies the condition of the lemma.

    \item [$\bullet$] Case (ii): $c$ is redundant with respect to $(\sigma,\delta)$.

    By definition of redundancy, there exists a subplan of $c$ which
    are equivalent to $c$ with respect to $(\sigma,\delta)$.
    Let $c'$ be a subplan of $c$ which are equivalent to $c$
    with respect to $(\sigma,\delta)$ whose length is minimal among all subplans
    which are equivalent to $c$ with respect to $(\sigma,\delta)$.  To prove the lemma,
    it is enough to show that $c'$ is not redundant with respect to $(\sigma,\delta)$.
    Assume the contrary, there exists a subplan $c''$ of $c'$ which
    is equivalent to $c$ with respect to $(\sigma,\delta)$. Trivially,
    the number of actions in $c''$ is smaller than the number
    of actions in $c'$. By definition, we have that $c''$ is also a subplan of $c$
    which equivalent
    to $c$ with respect to $(\sigma,\delta)$. This contradicts the fact that
    $c'$ has the minimal length among all subplans of $c$ which are equivalent to $c$.
     So, we conclude that
     $c'$ is not redundant with respect to $(\sigma,\delta)$. The lemma is
     proved. 
\end{itemize}
\end{proof}

\st
{\lemma \label{lemAdd3-son} Let $\sigma$ be an a-state and $c$ be a case plan
$c = a; case\ (\varphi_1 \rightarrow p_1
        \ldots \varphi_m \rightarrow p_m\ )$
such that $\bot \not \in \Phi^*(c,\sigma)$. Then, if $Sens_a
\setminus \sigma \ne \emptyset$, there exists a possibly
regressable plan $c' = a; case\ (\varphi_1' \rightarrow p'_1
        \ldots \varphi_n' \rightarrow p'_n\ )$
such that $\Phi^*(c,\sigma) = \Phi^*(c', \sigma)$.
}
\begin{proof} [Lemma \ref{lemAdd3-son}]
We prove the lemma by constructing $c'$. Let $S =
\{\varphi_1,\ldots,\varphi_m\}$ and $S_a = Sens_a \setminus
\sigma$. Let $L = \{f \mid f \in S_a\} \cup \{\neg f \mid f \in
S_a\}$. First, observe that because of $\bot \not \in
\Phi^*(c,\sigma)$ we have that $a$ is executable in $\sigma$ and
for each $\sigma' \in \Phi^*(c,\sigma)$ there exists one
$\varphi_i \in S$ such that $\varphi_i$ is satisfied in $\sigma'$.
Furthermore, without the lost of generality, we can assume that
for each $\varphi_i \in S$, there exists (at least) one $\sigma'
\in \Phi^*(c,\sigma)$ such that $\varphi_i$ is satisfied in
$\sigma'$.

\st It is easy to see that for each $i$, we can write $\varphi_i =
\psi_i \wedge \chi_i$ where $\psi_i$ is the conjunction of
literals occurring in $\varphi_i$ and belonging to $L$ and
$\chi_i$ is the conjunction of literals that do not belong to $L$.
From the above observation, we have that $\chi_i$ is satisfied by
$\sigma$. So, $\varphi = \wedge_{i=1}^m \chi_i$ holds in $\sigma$.
Thus, the conditional plan $c_1 = a; case\ (\varphi'_1 \rightarrow
p_1 \ldots \varphi'_m \rightarrow p_m\ )$ where $\varphi'_i =
\psi_i \wedge \varphi$ satisfies that
 $\Phi^*(c,\sigma) = \Phi^*(c_1, \sigma)$.

\st
Since $\psi_i$ is a consistent conjunction of literals from $L$
and $\psi_i$'s are mutual exclusive, there exists a partition
$(S_1,\ldots,S_m)$ of $BIN(S_a)$ such that
for every $\eta \in S_i$, $\eta = \psi_i \wedge \eta'$. Let \\
$c_2 = a; case\ (\gamma^1_1 \rightarrow p_1 \ldots \gamma^{|S_1|}_1 \rightarrow p_1 $ \\
\hspace*{0.8in} $\gamma^1_2 \rightarrow p_1 \ldots \gamma^{|S_2|}_2 \rightarrow p_2$ \\
\hspace*{0.8in} $ \ldots $ \\
\hspace*{0.8in} $\gamma^1_m \rightarrow p_1 \ldots \gamma^{|S_m|}_m \rightarrow p_m\ )$
\\
where $\gamma^j_i = \eta^j_i \wedge \varphi \wedge \gamma$, $S_i =
\{\eta^1_i,\ldots,\eta^{|S_i|}_i\}$ for $i=1,\ldots,m$, and
$\gamma = \wedge_{f \in Sens_a \cap \sigma.T} f \wedge
        \wedge_{f \in Sens_a \cap \sigma.F} \neg f$.
We have that $\Phi^*(c,\sigma) = \Phi^*(c_2, \sigma)$. It is easy
to see that the set $\{\gamma^1_1,\ldots,\gamma^{|S_m|}_m\}$ spans
over $S_a$ and $Sens_a \subseteq (\gamma^j_i)^+ \cup
(\gamma^j_i)^-$. Thus, $c_2$ is possibly regressable. The lemma is
proved with $c' = c_2$. 
\end{proof}

\setcounter{section}{3}

\setcounter{theorem}{21}

\st {\lemma \label{lemAdd3} Let $\sigma$ be an a-state, let
$\delta$ be a p-state, and let $c$ be a normalized conditional
plan such that $\bot \not \in \Phi^*(c,\sigma)$ and
$\Phi^*(c,\sigma) \subseteq ext(\delta)$. There exists a
normalized plan $c'$ such that $c'$ is regressable with respect to
$(\sigma,\delta)$ and $c'$ is equivalent to $c$ with respect to
$(\sigma,\delta)$.}

\begin{proof} [Lemma \ref{lemAdd3}] We will prove the lemma using
induction on $count(c)$, the number of case plans in $c$.

\begin{itemize}
\item [$\bullet$] Base case: $count(c) = 0$

    This follows from Lemma \ref{lemAdd2}. 
\item [$\bullet$] Inductive Step: Assume that we have proved the
    lemma for $count(c) \leq k$. We need to prove the lemma for
    $count(c)=k+1$.

    By construction of $c$, we have two cases
    \begin{enumerate}
    \item $c= a; p$ where
    $p= case\ (\varphi_1 \rightarrow p_1 \ldots \varphi_m \rightarrow p_m)$.
    Here, we have two cases.
    \begin{enumerate}
    \item $Sens_a \setminus \sigma = \emptyset$. In this case, we have
    that there exists some $j$ such that $\varphi_j$ is satisfied by $\sigma$
    and $\Phi^*(c, \sigma) = \Phi^*(p_j, \sigma)$.
    Thus, $c$ is equivalent to $p_j$ with respect to $(\sigma, \delta)$.
    Since $count(p_j) < count(c)$, by inductive hypothesis and
    the transitivity of the equivalence relation, we conclude that
    there exists a normalized plan $c'$ such that
    $c'$ is regressable with respect to $(\sigma,\delta)$ and $c'$
    is equivalent to $c$ with respect to $(\sigma,\delta)$.

    \item $Sens_a \setminus \sigma \ne \emptyset$.
    Using Lemma \ref{lemAdd3-son}, we can construct a normalized
    plan
$c_1 = a; case\ (\varphi_1' \rightarrow p_1' \ldots \varphi'_n \rightarrow
        p'_n)$
    which is possibly regressable and
    $\Phi^*(c_1,\sigma) =  \Phi^*(c, \sigma)$.
    From the construction of $c_1$, we know that for each
    $\sigma' \in \Phi(a,\sigma)$ there exists one and only one
    $j$, $1 \le j \le n$, such that $\varphi_i'$ is satisfied
    in $\sigma'$. Applying the inductive hypothesis
    for $(\sigma',\delta)$ and the plan $p_i'$,
    we know that there exists normalized regressable plan $q_i$
    which is equivalent to $p_i'$ with respect to $(\sigma',\delta)$.
    This implies that
    $c' = a; case\ (\varphi_1' \rightarrow q_1
        \ldots \varphi'_n \rightarrow q_n)$
    is equivalent to $c$ with respect to $(\sigma,\delta)$.
    Furthermore, every case plan in $c_2$ is possibly regressable
    and each $q_i$ is regressable with respect to $(\sigma',\delta)$.
    To complete the proof, we will show that $c'$ is not
    redundant with respect to $(\sigma,\delta)$. From the assumption
    that $Sens_a \setminus \sigma \ne \emptyset$ and the
    construction of $c'$, we know that we cannot replace
    $a$ with some $SubSense(a)$. Furthermore, because
    for each $\sigma' \in \Phi(a,\sigma)$ there exists
    at most one $j$ such that $\varphi_j'$ is satisfied in
    $\sigma'$, none of the branches can be removed.
    This,
    together with the fact that $q_j$ is not redundant with respect to
    $(\sigma',\delta)$, implies
    that $c'$ is not redundant with respect to $(\sigma,\delta)$.
    The inductive hypothesis is proved for this case as well.
    \end{enumerate}

    \item $c = \alpha;c_1$ where $\alpha$ is a sequence of
    non-sensing actions and $c_1$ is a case plan.
    Let $P_\alpha = \{\alpha' \mid \alpha'$ is a subplan
    of $\alpha$ and
    $\alpha';c_1$ is equivalent to $c$ with respect to $(\sigma,\delta)\}$.
    Let $\beta$ be a member of $P_\alpha$
    such that $|\beta| = \min \{|\alpha'| \mid \alpha' \in P_\alpha\}$.
    We have that $\beta$ is a sequence of non-sensing actions,
    and so, there exists only one a-state in
    $\Phi^*(\beta, \sigma)$.
    Let us denote the unique a-state in $\Phi^*(\beta, \sigma)$
    by $\sigma_1$. It follows from the above
    case and the inductive hypothesis that
    there exists a normalized, regressable plan $c_1'$ which
    is equivalent to $c_1$ with respect to $(\sigma_1,\delta)$.
    Consider the plan $c' = \beta; c_1'$.
    We have that $c'$ is a normalized, possibly regressable
    conditional plan. To complete the proof, we will show that
    $c'$ is not redundant with respect to $(\sigma,\delta)$.
    Assume the contrary, we will have three cases:
\begin{enumerate}
\item There exists a subplan $\beta'$ of $\beta$
    such that $q = \beta';c_1'$ is equivalent to $c'$
    with respect to $(\sigma,\delta)$. This implies that
    $\beta';c_1$  is equivalent to $c'$
    with respect to $(\sigma,\delta)$ which contradicts the
    construction of $\beta'$.
\item There exists a subplan $c''$ of $c_1'$
    such that $q = \beta;c''$ is equivalent to $c'$
    with respect to $(\sigma,\delta)$. This implies that
    $c''$  is equivalent to $c_1'$
    with respect to $(\sigma_1,\delta)$ which contradicts the
    construction of $c_1'$.
\item  There exists a subplan $\beta'$ of $\beta$ and
     a subplan $c''$ of $c_1'$
    such that $q = \beta';c''$ is equivalent to $c'$
    with respect to $(\sigma,\delta)$. Similar arguments
        as in the above cases allow us to conclude that this
    case cannot happen as well.
    \end{enumerate}
    This shows that $c'$ is not redundant with respect to $(\sigma,\delta)$.
    So, we have proved that $c'$ is normalized, regressable,
    and equivalent to $c$
    with respect to $(\sigma,\delta)$.
    The inductive step is proved for this case. 
    \end{enumerate}

    \end{itemize}

\end{proof}

\setcounter{section}{3}

\setcounter{theorem}{22}

\begin{theorem} [Completeness of Regression] \label{theoAdd1}
Given a planning problem $P= \langle A,O,I,G \rangle$ and a
progression solution $c$ of $P$. There exists a normalized
regression solution $c'$ of $P$ such that $c'$ is equivalent to
$c$ with respect to $(\sigma_I, \delta_G)$.
\end{theorem}

\begin{proof} [Theorem \ref{theoAdd1}] Let $c' = normalized(c)$. It
follows from Lemma \ref{lemAdd1} that $\Phi^*(c',\sigma_I) =
\Phi^*(c, \sigma_I)$. Lemma \ref{lemAdd3} implies that there
exists a normalized regressable plan $c''$ with respect to
$(\sigma_I, \delta_G)$ which is equivalent to $c$ with respect to
$(\sigma_I, \delta_G)$. The conclusion of the theorem follows
directly from Lemma \ref{lem31b} and Theorem \ref{lem3}.
\end{proof}

\setcounter{section}{4}

\setcounter{theorem}{0}

\begin{theorem} \label{theo41}
For every $\langle c, \delta \rangle \in N$ where $N$ denotes the
set of plan-state pairs maintained by {\bf Solve}$(P)$,
$Regress^*(c,\delta_G) = \delta$.
\end{theorem}
\begin{proof}  [Theorem \ref{theo41}]
Observe that {\bf Solve($P$)} adds one plan-state pair to $N$ per
iteration (Steps 2-5). Thus, for each element $\langle c, \delta
\rangle$ of $N$ there exists a number $l$ such that $\langle c,
\delta \rangle$ is added to $N$ during the $l^{th}$ iteration of
{\bf Solve($P$)}. We refer to $l$ as the {\em iteration number of}
$\langle c, \delta \rangle$ and denote it by $l_{\langle c, \delta
\rangle}$. We prove the theorem by induction on $l_{\langle c,
\delta \rangle}$ that $Regress^*(c,\delta_G) = \delta$.

\begin{itemize}
    \item [$\bullet$] Base case: $|l_{\langle c,\delta \rangle}| = 0$.
    This implies that $c =[]$ and $\delta = \delta_G$.
    Clearly, $Regress^*(c,\delta_G) = \delta_G$.
    The base case is proved.

    \item [$\bullet$] Inductive Step:
        Assume that we have proved the theorem for
    every plan-state pair $\langle c,\delta \rangle$ with
$|l_{\langle c,\delta \rangle}| \leq k$.
        We now show that the theorem is correct for $\langle c,\delta \rangle$
    with $|l_{\langle c,\delta \rangle}| = k+1$.
    We have two cases:

\begin{enumerate}
    \item There exists a plan-state pair $\langle c',\delta' \rangle \in N$
    such that $l_{\langle c',\delta' \rangle} \leq k$,
    $c = a;c'$, $a$ is a non-sensing action
    and $Regress(a,\delta') = \delta$. It follows from the
    inductive hypothesis that
    $Regress^*(c',\delta_G) = \delta'$. The conclusion of the inductive
    step  for this case follows from
        \[Regress^*(c, \delta_G) = Regress^*(a,Regress^*(c',\delta_G))
    = Regress(a,\delta') = \delta.\]

   \item There exists a set of plan-state pairs $
    \Delta = \{\langle c_1, \delta_1 \rangle, \ldots,
            \langle c_n, \delta_n \rangle\}  \subseteq N
    $, a set of formulas $\{\varphi_1,\ldots,\varphi_n\}$
and     a sensing action $a$
such that $l_{\langle c_i, \delta_i \rangle} \leq k$ for every $i$,
    $1 \le i \le n$, $a$, $\varphi$'s and $\Delta$ satisfy the conditions
    specified in Step 4.2 of {\bf Solve($P$)},
    $Regress(a, \{[\delta_i.T \cup \varphi_i^+,
            \delta_i.F \cup \varphi_i^-] \mid i=1,\ldots,n\}) = \delta$.
    Let
    \[c = a; case(\varphi_1 \rightarrow c_1, \ldots, \varphi_n \rightarrow
        c_n).\]
    By inductive hypothesis, we have that
    $Regress^*(c_i,\delta_G) = \delta_i$.
        Thus, the conclusion of the inductive
    step  for this case follows from
        \[Regress^*(c, \delta_G) = Regress(a,\{R(c_1,\delta_G),\ldots,R(c_n,\delta_G)\}) = \delta.\]

\end{enumerate}
\end{itemize}
\end{proof}

\setcounter{section}{8}

\setcounter{theorem}{11}

\begin{lemma} \label{lem41-son1}
Given a planning problem $P = \langle A,O,I,G \rangle$ and a
plan-state pair $\langle c_1,\delta_1\rangle$ belonging to $N$,
the set of plan-state pairs maintained by $\textbf{Solve}(P)$. Let
$c_2$ be a sequence of non-sensing actions and $\delta_2$ be a
p-state such that $Regress^*(c_2,\delta_1) = \delta_2 \ne \bot$.
Then, $N$ contains a plan-state pair $\langle c,\delta_2\rangle$.
\end{lemma}

\begin{proof} [Lemma \ref{lem41-son1}] We prove by induction on the length of $c_2$,
$|c_2|$.
\begin{itemize}
\item [$\bullet$] Base Case: $|c_2|=0$. Obvious since $\delta_2 =
\delta_1$ and  $\langle c_1,\delta_1\rangle \in N$.

\item [$\bullet$] Inductive step: Assume that we have proved the
lemma for $|c_2| \le k$. We need to prove it for $|c_2| = k+1$.
Let $c_2 = \beta; a$ where $|\beta| = k$. Let $Regress(a,\delta_1)
= \delta_3 $. By Definition \ref{reg-nonsensing}, we have that
$Regress^*(c_2,\delta_1) = Regress^*(\beta, Regress(a,\delta_1))$.
Because $\delta_2 \ne \bot$, we have that $\delta_3 \ne \bot$.
Since $\textbf{Solve}(P)$ repeats Steps 2-5 until $N_s$ does not
change, we conclude that either $\langle c_1',\delta_3\rangle \in
N$ where $c'_1 = c_1;a$ or there exists some node $\langle
c_1'',\delta_3\rangle \in N$. Applying the inductive hypothesis
for this plan-state pair and the sequence $\beta$ with
$Regress^*(\beta, \delta_3) = \delta_2$ we have that there exists
a node $\langle c,\delta_2\rangle$ in $N$. The inductive step is
proved.
\end{itemize}
\end{proof}

\begin{lemma} \label{lem41-son2}
Given a planning problem $P = \langle A,O,I,G \rangle$ and a
plan-state pair $\langle c_1,\delta_1\rangle$ belonging to $N$,
the set of plan-state pairs maintained by $\textbf{Solve}(P)$. Let
$c_2$ be a normalized plan and $\delta_2$ be a p-state such that
$Regress^*(c_2,\delta_1) = \delta_2 \ne \bot$. If there exists an
a-state $\sigma \in ext(\delta_2)$ and $c_2$ is regressable with
respect to $(\sigma ,\delta_1)$, then $N$ contains a plan-state
pair $\langle c,\delta_2\rangle$.
\end{lemma}

\begin{proof} [Lemma \ref{lem41-son2}] We prove the
lemma by induction on $count(c_2)$.
\begin{itemize}
\item [$\bullet$] Base Case: $count(c_2) = 0$. This follows
immediately from Lemma \ref{lem41-son1}.

\item [$\bullet$] Inductive Step: Assume that we have proved the
lemma for $|count(c_2)| \le k$. We need to prove it for
$|count(c_2)| = k+1$. It is easy to see that we can assume that
$c_2 = \alpha; p$ where $p = a; case(\varphi_1 \rightarrow p_1,
\ldots, \varphi_n \rightarrow p_n)$ where $\alpha$ is a sequence
of non-sensing actions, $p_i$ are normalized plans, and
$\{\varphi_1,\ldots,\varphi_n\}$ spans over a set $\emptyset \ne
S_a \subseteq Sens_a$. We have that $count(p_i) \le k$.

\st Let $\gamma_i = Regress^*(p_i, \delta_1)$ for $i = 1, \ldots,
n$. Since $Regress^*(c_2,\delta_1) = \delta_2 \ne \bot$, we have
that $\gamma_i \ne \bot$ ($i = 1, \ldots, n$). Let $\Phi^*(\alpha,
\sigma) = \{\sigma_1\}$, it follows from the fact that $c_2$ is
regressable with respect to $(\sigma,\delta_1)$ and Lemma
\ref{lem31b-son} that for each $p_i$ ($1 \leq i \leq n$), there
exists a unique $\sigma' \in \Phi(a, \sigma_1)$ such that $p_i$ is
regressable with respect to $(\sigma',\delta_1)$; and by Lemma
\ref{lem31b} we have that $\sigma' \in ext(\gamma_i)$. From the
inductive hypothesis, we conclude that there exist conditional
plans $q_i$, $1 \le i \le n$, such that $\langle q_i, \gamma_i
\rangle$ belong to $N$ where $\gamma_i = Regress^*(q_i, \delta_G)$
(by Theorem \ref{theo41}).

\st
Since $Regress^*(c_2, \delta_1) \ne \bot$
we conclude that
\[
Regress(a, \{R(p_1,\delta_1), \ldots, R(p_n,\delta_1)\}) = \delta'
\ne \bot.
\]

\st Because $c_2$ is a regressable plan with respect to
$(\sigma,\delta_1)$, we know that $\{\varphi_1,\ldots,\varphi_n\}$
spans over a set $S_a$, $\emptyset \ne S_a \subseteq Sens_a$.
Since $\delta' \neq \bot$ we have that $a$ is applicable in
$\Delta = \{R(p_1,\delta_1), \ldots, R(p_n,\delta_1)\}$. Now,
consider the case plan $p' = a; case(\varphi_1 \rightarrow q_1,
\ldots, \varphi_n \rightarrow q_n)$. We have that $\gamma_i =
Regress^*(q_i, \delta_G)$ ($i=1, \ldots, n$) and $a$ is applicable
in $\Delta$, it follows from Step 4.2 of $\textbf{Solve}(P)$,
there exists a plan-state pair $\langle c', \delta' \rangle$ in
$N$. Because $Regress^*(\alpha, \delta') = \delta_2$ and $\alpha$
is a sequence of non-sensing actions, Lemma \ref{lem41-son1}
implies that $N$ contains some plan-state pair $\langle c,
\delta_2 \rangle$.
\end{itemize}
\end{proof}

\setcounter{section}{4}

\setcounter{theorem}{1}

\begin{theorem} \label{theo42}
For every planning problem $P = \langle A,O,I,G \rangle$,
\begin{enumerate}
\item {\bf Solve}$(P)$ will always stop;
\item if $P$ has a regression solution $c$ then {\bf Solve}$(P)$ will return a
conditional plan $c'$ such that $Regress(c,\delta_G) =
Regress(c',\delta_G)$; and
\item if $P$ has no regression solution
then {\bf Solve}$(P)$ will return NO SOLUTION.
\end{enumerate}
\end{theorem}

\begin{proof} [Theorem \ref{theo42}] We will prove (2) using Lemma
\ref{lem41-son2} and (3) using (2).

\begin{enumerate}
        \item
        Since we only consider domains with finite number of actions and fluents,
        the set of p-states is finite.
    Given a set of plan-state pairs $N$, $\textbf{Solve}(P)$
    either adds a new plan-state pair $\langle c, \delta \rangle$
    to $N$ in the step 4, where $\delta \not\in N_s$ or stop.
    Because the set of p-states is finite, we can conclude
    that $\textbf{Solve}(P)$ will eventually terminate.

        \item If $P$ has a regression solution:
    let $c$ be a regression solution of $P$.
    We have that $\Phi^*(c, \sigma_I) \subseteq ext(\delta_G)$. It
    follows from Lemmas \ref{lemAdd1} and \ref{lemAdd3} that there exists a normalized
    plan $c'$ that is equivalent to $c$ and regressable with respect
    to $(\sigma_I,\delta_G)$.
    Applying Lemma \ref{lem41-son2} for
    the plan-state pair $\langle [],\delta_G \rangle$,
    the plan $c'$, and the p-state $\delta = Regress^*(c',\delta_G)$
    (we have $\sigma_I \in ext(\delta)$ by Lemma \ref{lem31b}),
    we can conclude that there exists a
    plan-state pair $\langle p, \delta \rangle$ in $N$,
    the set of plan-state pairs maintained by $\textbf{Solve}(P)$.
    It follows from Theorem \ref{theo41}
    that $Regress^*(p, \delta_G) = \delta$.
        Since $\sigma_I \in ext(\delta)$, the step 3 of the algorithm will
        return $p$ as a regression solution.

        \item If $P$ has no a regression solution: From (1),
        we have that
    $N$ is finite and the algorithm will eventually stop.
        If $P$ has a solution then it will return one at step 3.
        Since $P$ has no solution and $N$ is finite,
        the algorithm will eventually go to step 6, i.e. it will
        return NO SOLUTION.

    \st From cases (1), (2), and (3), the theorem is proved.

\end{enumerate}
\end{proof}

\begin{acks}
We are grateful to the anonymous referees whose useful comments
helped us to improve this paper. This work was supported by NSF
grant number 0070463 and NASA grant number NCC2-1232.
Tran Cao Son was 
supported by 
EIA-0220590.
\end{acks}

\bibliographystyle{acmtrans}

\begin{received}
Received Month Year;

revised Month Year; accepted Month Year
\end{received}

\end{document}